\documentclass[11pt]{article}

\usepackage[preprint]{acl}

\usepackage{times}
\usepackage{latexsym}

\usepackage[T1]{fontenc}

\usepackage[utf8]{inputenc}

\usepackage{microtype}

\usepackage{inconsolata}

\usepackage{graphicx}

\usepackage{xcolor}
\usepackage{enumitem}
\usepackage{amsmath}
\usepackage{amsfonts}
\usepackage{booktabs}
\usepackage{multirow}
\usepackage{changepage}
\usepackage{array}

\newcommand{\ourmethod}[0]{{CLiMRS}}
\newcommand{\ourbenchmark}[0]{{CLiMBench}}

\title{Leveraging Adaptive Group Negotiation for Heterogeneous \\
       Multi-Robot Collaboration with Large Language Models}

\author{
  \textbf{Siqi Song}$^{1*}$ \ 
  \textbf{Xuanbing Xie}$^{2*}$ \ 
  \textbf{Zonglin Li}$^{3*}$ \ 
  \textbf{Yuqiang Li}$^{3}$ \ 
  \textbf{Shijie Wang}$^{3\dagger}$ \ 
  \textbf{Biqing Qi}$^{3\dagger}$ \ 
  \\
  \\
  $^{1}$ Tsinghua University \quad 
  $^{2}$ Central South University \quad
  $^{3}$ Shanghai AI Laboratory
  \\
  {\fontsize{10.5pt}{11pt}\selectfont
    $^{*}$ Equal Contribution.
    $^{\dagger}$ Equal Advising.
    Correspondence: \href{mailto:songsq21@mails.tsinghua.edu.cn}{songsq21@mails.tsinghua.edu.cn}
  }
}

\begin{document}
\maketitle

\begin{abstract}

Multi-robot collaboration tasks often require heterogeneous robots to work together over long horizons under spatial constraints and environmental uncertainties.
Although Large Language Models (LLMs) excel at reasoning and planning, their potential for 
coordinated control has not been fully explored.
Inspired by human teamwork,
we present \ourmethod{}
(\underline{\textbf{C}}ooperative Large-\underline{\textbf{L}}anguage-Model-Dr\underline{\textbf{i}}ven Heterogeneous \underline{\textbf{M}}ulti-\underline{\textbf{R}}obot \underline{\textbf{S}}ystem)
, an adaptive group negotiation framework among LLMs for multi-robot collaboration.
This framework pairs each robot with an LLM agent and dynamically forms subgroups through a general proposal planner.
Within each subgroup, a subgroup manager leads perception-driven multi-LLM discussions to get commands for actions.
Feedback is provided by both robot execution outcomes and environment changes.
This grouping–planning–execution–feedback loop enables efficient planning and robust execution.
To evaluate these capabilities, we introduce \ourbenchmark{}, a heterogeneous multi-robot benchmark of challenging assembly tasks.
Our experiments show that \ourmethod{} surpasses the best baseline, achieving over 40\% higher efficiency on complex tasks without sacrificing success on simpler ones.
Our results demonstrate that leveraging human-inspired group formation and negotiation principles markedly enhances the efficiency of heterogeneous multi-robot collaboration.
Our code is available \href{https://github.com/song-siqi/CLiMRS}{\nolinkurl{here}}.

\end{abstract}


\section{Introduction}

Addressing real-world, everyday tasks often requires collaboration to efficiently handle long-horizon planning and complex perception.
However, developing embodied agents for multi-robot tasks remains an open challenge.
Inspired by human teamwork, incorporating human teaming principles into multi-robot collaboration, where groups of robotic agents coordinate planning and perception through shared observations and information, offers a promising yet challenging path to improving efficiency and robustness \cite{zhang2024building}.

Meanwhile, large language models (LLMs) have demonstrated strong capabilities across a range of tasks, including question answering~\cite{rein2024gpqa}, code generation~\cite{jain2024livecodebench}, and logical reasoning~\cite{plaat2024reasoning}.
Recent work has integrated LLMs into robotic planning~\cite{song2023llm, zhang2024badrobot, mower2024ros, salimpour2025towards, liang2025large}, with several studies extending these approaches to multi-robot collaboration~\cite{zhang2024building, mandi2024roco, liu2025coherent}.
However, previous work has mainly focused on homogeneous agents~\cite{liu2024leveraging}, limiting the diversity of capabilities that can be exhibited during collaboration.
Moreover, existing work on heterogeneous teams often assumes idealized operating conditions~\cite{liu2025coherent}, overlooking cumulative errors over long horizons, underestimating communication costs, and overestimating cooperative efficiency.
As a result, despite the promise of LLM-driven multi-robot collaboration, significant gaps remain under heterogeneous agents, long-horizon objectives, and real-world constraints.

\begin{figure*}[t]
  \centering
   \includegraphics[width=1.0\linewidth]{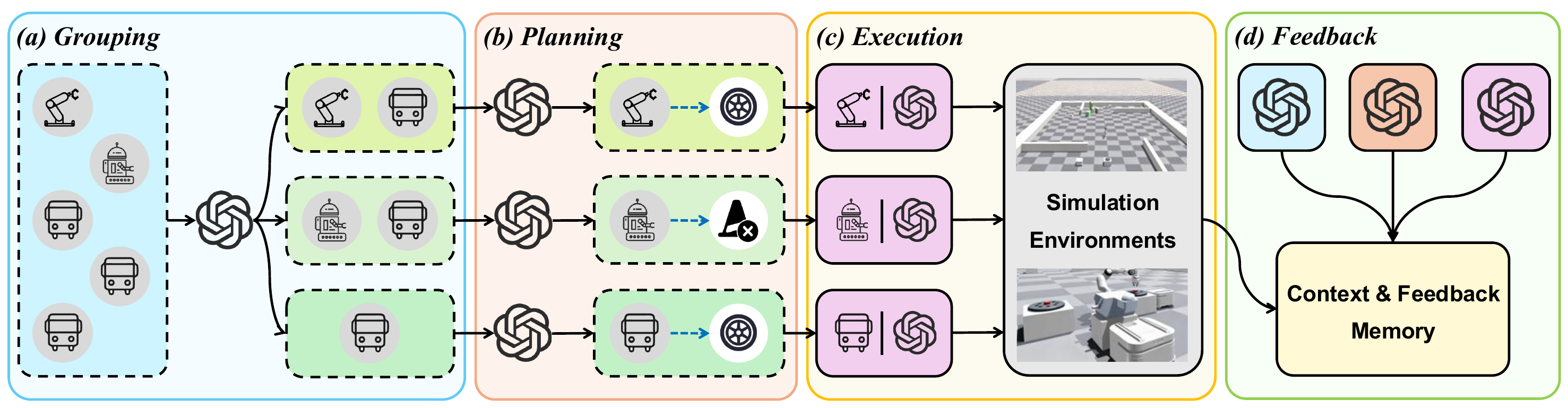}
   \caption{\textbf{Overview.}
    We present \ourmethod{}, an adaptive group negotiation framework among multiple LLMs that enables multi-robot collaboration through a grouping–planning–execution–feedback loop, and \ourbenchmark{}, a heterogeneous multi-robot collaboration benchmark in simulation with challenging long-horizon assembly tasks.
}
   \label{fig:climrs}
\end{figure*}

To address these limitations, we propose \textbf{\ourmethod{}} 
(\underline{\textbf{C}}ooperative Large-\underline{\textbf{L}}anguage-Model-Dr\underline{\textbf{i}}ven Heterogeneous \underline{\textbf{M}}ulti-\underline{\textbf{R}}obot \underline{\textbf{S}}ystem), 
an adaptive group negotiation framework among multiple LLMs that enables multi-robot collaboration.
\ourmethod{} orchestrates heterogeneous robots via dynamic subgroup formation and cooperative planning, enabling robust long-horizon collaboration in uncertain environments.
Within this framework, each robot is paired with an independent LLM agent that communicates with its peers to accomplish complex, long-horizon tasks. To enhance collaborative effectiveness, the system leverages the broad world knowledge of LLMs and explicitly models inter-agent dependencies through a grouping–planning–execution–feedback loop.

To systematically evaluate the applicability of \ourmethod{} in challenging scenarios where heterogeneous robots must cope with unpredictable execution errors, we introduce \textbf{\ourbenchmark{}}, a benchmark for heterogeneous multi-robot collaboration.
\ourbenchmark{} is built around long-horizon assembly tasks that require robots to jointly perform object search, navigation, transportation, and assembly under partial observability.
It features five robotic devices across three types of heterogeneous robots.
Tasks of varying difficulty simulate material-handling and assembly processes with diverse skill usage, designed to test the planning and perception capabilities of LLM-based frameworks.

We evaluated our framework in \ourbenchmark{} and another heterogeneous robot collaboration benchmark~\cite{liu2025coherent}.
Our experiments show that \ourmethod{} outperforms the best baseline, increasing success rates and improving efficiency on complex tasks while maintaining high success on simpler ones.
These results demonstrate that incorporating human-inspired dynamic subgroup formation and negotiation principles substantially enhances the efficiency of heterogeneous multi-robot collaboration.
To summarize, our main contributions are:
\begin{itemize}
\setlength{\itemindent}{0pt}
\setlength{\leftskip}{-10pt}
\setlength{\itemsep}{0em}
    \item We present \ourmethod{}, a multi-LLM cooperation framework for heterogeneous multi-robot collaboration that can perform long-horizon planning and efficient perception in complex tasks.
    \item We propose \ourbenchmark{}, a benchmark evaluating heterogeneous multi-robot collaboration with long-horizon assembly tasks, featuring varied skills and a realistic simulation environment.
    \item We demonstrate through extensive experiments that \ourmethod{} achieves significant efficiency improvements via dynamic group formation and cooperative long-horizon planning.
\end{itemize}
\section{Related Works}
\label{sec:headings}

\subsection{Robotic Skills Training Across Scenarios}
\textbf{Single Agent Skill Training.}
Current approaches to train embodied skills for task execution generally follow two primary paradigms: rule-based and learning-driven methods.
Traditional embodiment controllers optimize joint movements through the resolution of robotic kinematics, aiming to improve motion robustness and generate smoother, more precise trajectories~\cite{kashyap2021particle,katayama2023model}.
In recent years, advances in reinforcement learning and imitation learning have significantly improved robotic motion control across domains such as dexterous manipulation, bipedal locomotion, and quadrupedal navigation~\cite{rajeswaran2017learning, li2025reinforcement, bellegarda2024visual}.
These advances have progressively enabled embodied systems to coordinate actions in a cerebellum-like manner, supporting increasingly complex tasks in diverse environments.

\textbf{Multi-agent Skill Training.}
Originally developed in game AI~\cite{kurach2020google, perolat2022mastering}, multi-agent skill training has since been extended to more physically grounded domains such as robotics~\cite{wang2024multi, lai2025roboballet} and autonomous driving~\cite{li2022v2x}.
Despite these advances, multi-agent embodied learning remains limited by exponential state-space growth, and existing approaches such as mean-field methods~\cite{yang2018mean} struggle to generalize across heterogeneous robots in collaboration.


To further this goal, we design a set of diverse robotic skills in \ourbenchmark{} that model robots’ low-level execution outcomes, capturing both successful executions and failure cases, thereby providing a testbed for planning methods in heterogeneous multi-agent collaboration.

\begin{figure*}
  \centering
   \includegraphics[width=1.0\linewidth]{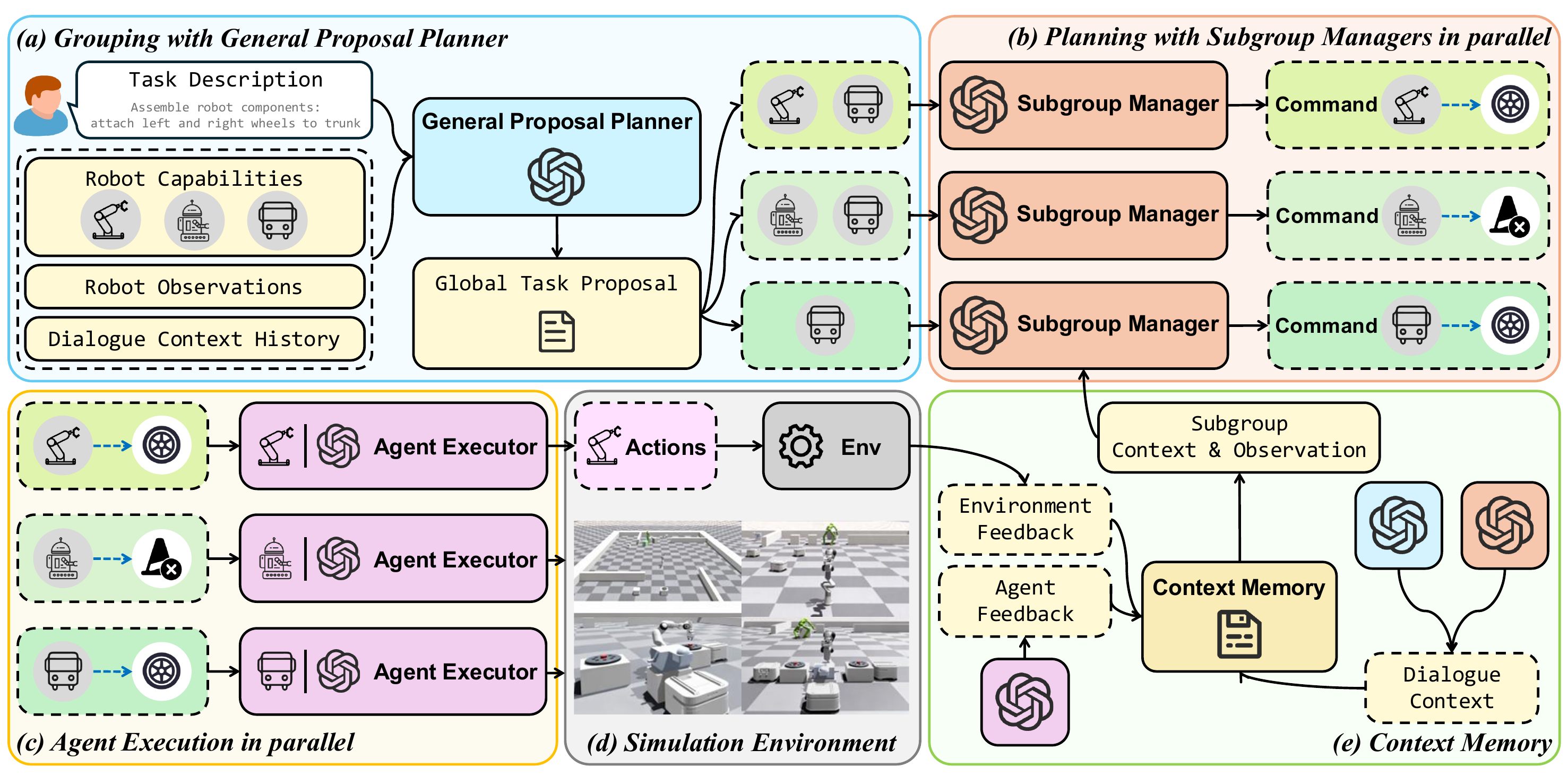}
   \caption{\textbf{\ourmethod{} Framework.}
   To employ our grouping–planning–execution–feedback cycle, \ourmethod{} comprises
   (a) a \emph{general proposal planner} that forms dynamic agent subgroups,  
   (b) multiple \emph{subgroup managers} for local agent commands,  
   (c) multiple \emph{agent executors} for robot skills and agent feedback,  
   (d) a \emph{simulation environment} for environment interaction and feedback, and  
   (e) a \emph{context memory} module for all dialogue context and feedback.
}
   \label{fig:method}
\end{figure*}

\subsection{Task Planning with LLMs in Robotics}

\textbf{LLM Planner for Robotics.}
The rapid progress of LLMs has motivated their use as task planners for robots, as few-shot and zero-shot learning enable the decomposition of high-level goals into executable task sequences~\cite{brown2020fewshot,huang2022language,llmzeroreason}.
Their code-generation abilities further allow LLMs to produce executable skill-level commands for robotic control~\cite{codeaspolicies,progprompt,wang2023voyager,wu2023tidybot}, while value-function-based methods guide robust skill selection~\cite{text2motion,saycan}.
Recent advances in prompting strategies further improve long-horizon LLM-based planning~\cite{zhang2024building,liu2025coherent}.
Some studies also explore reasoning in joint or 3D action spaces~\cite{mandi2023roco}.

\textbf{Multi-LLM Task Planning.}
Prior work emphasizes multi-LLM cooperation through discussion, debate, and role assignment~\cite{chen2023reconcile, liang2023encouraging, hong2024metagpt} to provide complementary perspectives and improve output reliability, often augmented with feedback and memory mechanisms to improve long-horizon planning~\cite{mandi2023roco, liu2025coherent, zhang2024building, wang2023voyager}.


\textbf{Decision-making Paradigms in Multi-Robot Collaboration.}
Two primary paradigms emerged for complex multi-robot tasks: centralized and decentralized approaches.
In decentralized schemes, multiple models or agents communicate, exchange intermediate plans, and iteratively refine their decisions through structured dialogue~\cite{mandi2023roco,zhang2024building,liu2024leveraginglargelanguagemodel},while centralized methods typically rely on a single LLM to decompose global objectives and allocate tasks when planning~\cite{kannan2023smart,liu2025coherent}.
A recent comparative study conducted in four diverse multi-agent scenarios~\cite{chen2023scalable} further reports that centralized communication consistently achieves higher success rates and markedly greater token efficiency with proper feedback mechanisms, highlighting its strong potential for scalable real-world deployment.

Based on previous findings, \ourmethod{} structures multi-LLM cooperation through a grouping–planning–execution–feedback loop. By dynamically forming subgroups for subtasks, the framework enables parallel planning, agent execution, and robust collaboration.

\section{Method}
\label{sec:method}


In this section, we present \ourmethod{}, an adaptive group negotiation framework among multiple LLMs that enables multi-robot collaboration.
In human team problem-solving, a common and effective strategy for tackling complex tasks is to decompose the problem and assign subtasks to subgroups that work in parallel.
Inspired by this, our approach forms dynamic agent subgroups that hold centralized discussions on robot perception in parallel, with each robot paired with an independent LLM agent to give feedback to discussions, thus forming a dynamic grouping–planning–execution–feedback cycle, which is shown in Fig.~\ref{fig:climrs}.

As illustrated in Fig.~\ref{fig:method},
\ourmethod{} comprises five core modules:  
(a) a \emph{general proposal planner} that forms dynamic agent groups,  
(b) multiple \emph{subgroup managers} that generate local agent commands,  
(c) multiple \emph{agent executors} that produce robot skills and return agent-level execution feedback,  
(d) a \emph{simulation environment} for real-time interaction and provides environment feedback, and  
(e) a \emph{context memory} module that records all inter-agent dialogues and feedback.

\subsection{Grouping with General Proposal Planner}
\label{sec:method_grouping}

The first stage of our cycle dynamically partitions agents into subgroups responsible for different aspects of the overall task, using a \emph{general proposal planner} to orchestrate the grouping process.

\textbf{General Proposal Planner.}
As illustrated in Fig.~\ref{fig:method}(a), the \emph{general proposal planner} generates a global task proposal that clusters agents into subtask-oriented teams.
Given the overall task instruction, the prompted LLM incorporates robot capabilities, current observations, and dialogue history through a structured prompt.
It outputs a plan with the following components:
(1) \emph{Situation Analysis}, assessing the environment and task progress;
(2) \emph{Spatial Analysis}, accounting for agent and object locations and spatial constraints;
(3) \emph{Task Decomposition}, breaking the objective into executable subtasks;
(4) \emph{Grouping Strategy}, clustering agents for parallel execution while minimizing interference;
(5) \emph{Subgoal Assignment}, specifying each group’s objective;
(6) \emph{Coordination Strategy}, outlining inter-group synchronization and execution order; and
(7) \emph{Risk Assessment}, identifying potential conflicts and mitigation plans.
The resulting group-to-subtask assignments are then passed to the perception and execution modules.


\subsection{Planning with Local Subgroup Managers}
\label{sec:method_planning}

Given the agent groupings and their assigned subtasks, the second stage generates precise robot-level commands based on robot capabilities and current observations.
As the subtasks are independent, multiple \emph{subgroup managers} operate in parallel, as illustrated in Fig.~\ref{fig:method}(b).

\textbf{Subgroup Manager.}
Each \emph{subgroup manager} conducts a centralized discussion within its subgroup to determine fine-grained commands for individual robots.
Conditioned on subtask instructions, subgroup robot capabilities, subgroup observations, and dialogue history, the subgroup manager selects appropriate skills and parameters for each agent.

\subsection{Execution and Environment Interaction}
\label{sec:method_execution_feedback}
With commands issued to robots, the final two stages of our cycle require them to evaluate these commands, determine appropriate actions, execute safely and provide feedback to refine future planning.  
The \emph{agent executor} LLM verifies the feasibility of its command and issues the corresponding action only when the command is deemed executable.  

\textbf{Agent Execution.}
Shown in Fig.~\ref{fig:method}(c), the \emph{agent executors} verify and execute commands from the \emph{subgroup manager} while providing feedback.
Each \emph{agent executor} LLM considers its robot’s capabilities, current observations, and available actions.
The executor first checks feasibility against the robot’s physical constraints and conditions.
If feasible, the action is executed using the robot’s skills; otherwise, the robot remains idle for this cycle.

\textbf{Simulation Environment.}
Shown in Fig.~\ref{fig:method}(d), the simulation environment serves as the execution backbone of our framework.
It receives the robot skill execution signals issued by the \emph{agent executors} and immediately carries out the corresponding low-level actions in real time.  
During execution, it monitors the evolving state of the environment and produces both updated robot observations and environment-level feedback.  
These outputs are fed back to the \emph{context memory} shown in Fig.~\ref{fig:method}(e), allowing the overall method to track task progress, refine its environment understanding, and supply the information required for the next round of the grouping–planning–execution–feedback cycle.

\begin{figure*}
  \centering
   \includegraphics[width=0.95\linewidth]{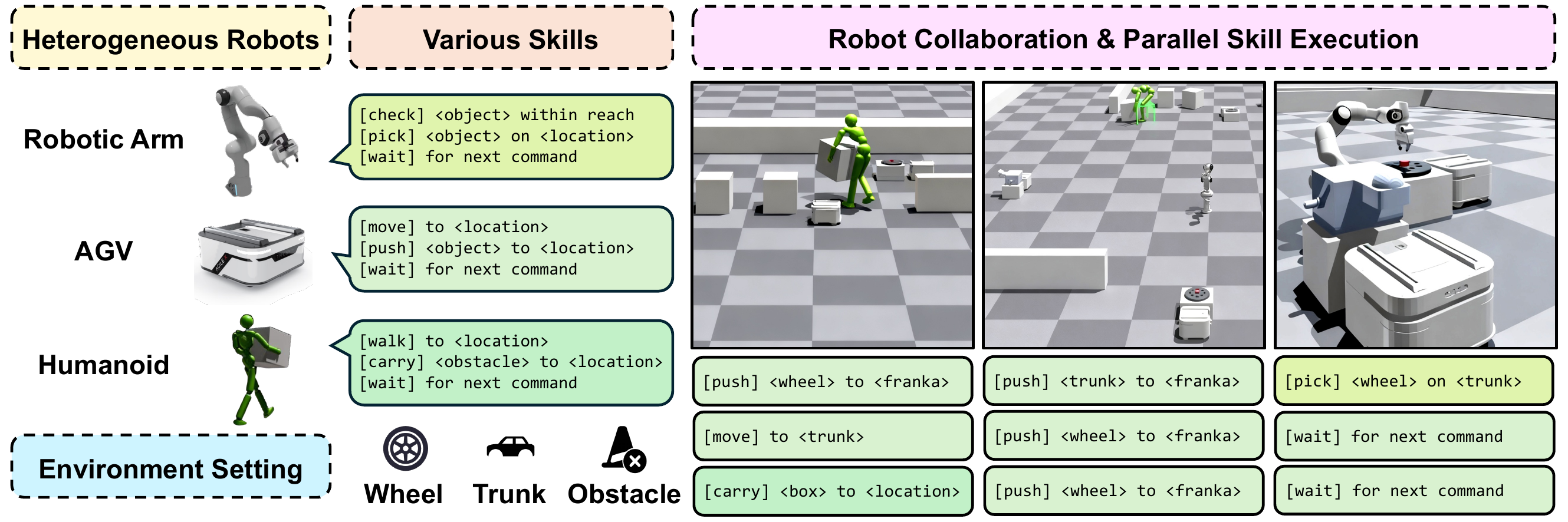}
   \caption{\textbf{Our Benchmark.}
   \ourbenchmark{} is a heterogeneous multi-robot collaboration benchmark, featuring multi-agent robots with diverse skills, enabling collaboration on complex assembly tasks of varying difficulty levels. 
}
   \label{fig:benchmark}
\end{figure*}

\subsection{Context Memory}
\label{method_mem_env}

With the grouping–planning–execution–feedback cycle described above, 
our framework employs a \emph{context memory} module that contains execution feedback, environment observations, and dialogue history, providing context to all LLM agents.

\textbf{Feedback Formation.}
Execution feedback is collected from two sources:
(1) \emph{environment feedback}, including updated environmental observations after robot actions are executed in the simulator, and
(2) \emph{agent feedback} generated by the \emph{agent executors}.
Agent executors report execution outcomes and provide diagnostic explanations for infeasible commands, enabling subsequent planners to reason about failures and task progress.



\textbf{Context Memory Organization.}
As illustrated in Fig.~\ref{fig:method}(d), the \emph{context memory} module stores:
(1) \emph{agent feedback}, including execution outcomes and diagnostic explanations from the \emph{agent executors};
(2) \emph{environment feedback}, consisting of updated observations from the \emph{simulation environment} after action execution; and
(3) \emph{dialogue context}, including current planning dialogue and dialogue history accumulated across previous cycles.

The context memory provides tailored context inputs for different modules.
For the \emph{general proposal planner}, it retains the most recent five dialogue turns along with the latest environment observations, enabling agent feedback to inform new task proposals and subgroup formations.
For the \emph{subgroup managers}, it stores each subgroup’s latest observations and the last five dialogue turns, supplying rich, localized context to guide fine-grained planning and perception within each subgroup.


\section{Benchmark}
\label{sec:benchmark}

In this section, we introduce \ourbenchmark{}, a benchmark for heterogeneous multi-robot collaboration.
We consider an assembly-oriented setting in which multiple components must be collected and assembled into a wheeled robot within a large workspace.
Unlike prior multi-agent collaboration benchmarks~\cite{liu2025coherent}, which decouple robot skill execution from the planning loop and assume successful execution by default, \ourbenchmark{} executes every robot skill within a realistic physics simulator, IsaacGym~\cite{makoviychuk2021isaac}.
This design explicitly allows execution failures and enables systematic evaluation of planning modules under realistic execution uncertainty.
The goal of \ourbenchmark{} is to evaluate long-horizon heterogeneous multi-robot collaboration in settings where planning and execution are tightly coupled.

\subsection{Task Setting and Scene Construction}

\textbf{Task Setting.}
The goal of the task is to assemble the components into a wheeled robot.
Within the workspace,
the components are distributed across different rooms, some of which are occluded or blocked by obstacles that must be cleared before access is possible.
As a result, robots must search for components, navigate complex environments, and perform coordinated manipulation under partial observability and execution uncertainty.
This long-horizon task is considered successfully complete once all required components are correctly assembled into the target product within a step count limit to evaluate task execution efficiency.

\textbf{Scene Construction.}
We instantiate this setting in the IsaacGym simulator by constructing the scene with multiple robotic agents and modular components, as is shown in Fig.~\ref{fig:benchmark}.
The environment includes three types of heterogeneous robots: one robotic arm, three autonomous ground vehicles (AGV), and one humanoid robot, each providing complementary capabilities for manipulation, transportation, and interaction with the environment to complete the assembly task.
To decouple high-level planning from low-level control while keeping heterogeneous robot capabilities, robot actions are executed through predefined skills, allowing high-level planners to reason over realistic execution outcomes while abstracting low-level control.

\textbf{Scene Initialization and Randomization.}
We initialize the environment and introduce controlled variations in task parameters and object configurations to enhance generalization.
At the start of each episode, robots execute their skills under randomized task conditions, leading to diverse skill sequences and varying levels of agent-wise interaction.
This provides a robust testbed for evaluating the effectiveness of different frameworks in multi-agent collaborative tasks.

\textbf{Environment feedback.}
We design the environment feedback along two dimensions:  
(1) We update the state of all agents and the coordinates of objects within their perceptible range, and  
(2) We report conflicts that arise when multiple robots execute skills simultaneously.  

\begin{table}[t]
\centering
\resizebox{1.0\linewidth}{!}{


\begin{tabular}{l >{\raggedright\arraybackslash}p{0.7\linewidth}}
\toprule
Robot type & Skill list \\
\midrule
Robotic Arm &
{[}check{]} <\emph{object}> within reach \\
&
{[}pick{]} <\emph{object}> on <\emph{location}> \\
&
{[}wait{]} for next command \\
\midrule
AGV &
{[}move{]} to <\emph{location}> \\
&
{[}push{]} <\emph{object}> to <\emph{location}> \\
&
{[}wait{]} for next command \\
\midrule
Humanoid &
{[}walk{]} to <\emph{location}> \\
&
{[}carry{]} <\emph{obstacle}> with both hands  \\
&
{[}wait{]} for next command \\
\bottomrule
\end{tabular}

}

\caption{
    \textbf{Robot Skills in \ourbenchmark.}  
    We assign a distinct set of skills for each robot based on its specific capabilities in the assembly tasks in \ourbenchmark{}.
}
\label{tab:skill_list}

\end{table}

\subsection{Robot Skill Design in \ourbenchmark{}}
\label{sec:robot-skill-execution}

In \ourbenchmark{}, each robot receives both the global task objectives and its observations (e.g., a humanoid robot observes its joint states, torso status, and target positions).
As is shown in Table~\ref{tab:skill_list}, we design distinct skills for different types of robots.
Further details are provided in the Appendix.

\textbf{Robotic Arm Manipulation.}
We use a Franka Panda arm to instantiate robotic manipulation skills in \ourbenchmark{}.
To balance speed and precision, we adopt a two-stage strategy consisting of a fast coarse motion followed by fine adjustment phase.
The arm is controlled using an operational-space controller (OSC) with gravity compensation, producing compliant spring--damper behavior in task space~\cite{narang2022factoryfastcontactrobotic}.
Smooth operational-space waypoints are generated via interpolation to ensure reliable execution.

\textbf{AGV Transportation.}
We use a TRACER Mini robot to instantiate AGV transportation skills in \ourbenchmark{}.
Navigation paths are planned using a Rapidly-exploring Random Tree (RRT) planner that accounts for environmental obstacles.
The resulting paths are executed via differential-drive control, enabling smooth turns during navigation.
For final delivery, the AGV follows straight-line motion to ensure reliable transportation and accurate placement at the assembly location.

\textbf{Humanoid Carrying.}
We instantiate humanoid carrying as a goal-conditioned control skill trained with Adversarial Motion Priors (AMP)~\cite{Peng_2021} following the single-object manipulation paradigm in COOHOI~\cite{gao2024coohoi}.
Given a target object and a goal location, the policy produces whole-body locomotion and manipulation to 
(1) approach and grasp the object, 
(2) carry it while maintaining balance and collision-robust gait, and 
(3) place it at the goal.
Training combines task rewards for reaching, holding, and placing, and AMP style rewards to encourage natural, stable motions and rapid recovery, yielding reliable execution under contact and minor disturbances.

\section{Experiments}
\label{sec:exp}

\begin{table*}[t]
\centering
\resizebox{0.7\linewidth}{!}{
\begin{tabular}{lcccccccc}
\toprule
\multirow{2}{*}{Method} &
\multicolumn{2}{c}{Mono-type Task} &
\multicolumn{2}{c}{Dual-type Task} &
\multicolumn{2}{c}{Trio-type Task} &
\multicolumn{2}{c}{Average} \\
\cmidrule(lr){2-3}\cmidrule(lr){4-5}\cmidrule(lr){6-7}\cmidrule(lr){8-9}
 & SR & AS & SR & AS & SR & AS & SR & AS \\
\midrule
DMRS-1D                  & 0.700 & 10.6 & 0.467 & 18.0 & 0.667 & 20.7 & 0.600 & 17.2 \\
DMRS-2D                  & 0.500 & 11.5 & 0.267 & 19.9 & 0.400 & 24.5 & 0.375 & 19.6 \\
CMRS                     & \textbf{0.900} & 7.9  & 0.533 & 16.4 & 0.533 & 22.2 & 0.625 & 16.5 \\
Primitive MCTS           & 0.000 & 14.0 & 0.000 & 21.5 & 0.000 & 26.9 & 0.000 & 21.7 \\
LLM-MCTS                 & 0.700 & 10.2 & 0.067 & 20.9 & 0.000 & 26.9 & 0.200 & 20.5 \\
COHERENT                 & \textbf{0.900} & 7.4 & \textbf{1.000} & 11.9 & \textbf{1.000} & 16.1 & \textbf{0.975} & 12.4 \\
\midrule
\ourmethod (Ours)        & \textbf{0.900} & \textbf{6.8} & \textbf{1.000} & \textbf{11.5} & \textbf{1.000} & \textbf{13.1} & \textbf{0.975} & \textbf{10.9} \\
Ground Truth (GT)        &   --  & 6.5  &   --  & 10.3 &   --  & 12.9 &   --  & 10.3 \\
\bottomrule
\end{tabular}
}
\caption{
    \textbf{Comparison Across Task Types in the COHERENT Benchmark.}
    \ourmethod{} outperforms all the baselines, achieving the largest gain on the most challenging trio-type tasks.\\
}
\label{tab:coherent1}
\end{table*}

\begin{table*}[t]
\centering
\resizebox{0.9\linewidth}{!}{
\begin{tabular}{lcccccccccccc}
\toprule
\multirow{2}{*}{Method} &
\multicolumn{2}{c}{S1} &
\multicolumn{2}{c}{S2} &
\multicolumn{2}{c}{S3} &
\multicolumn{2}{c}{S4} &
\multicolumn{2}{c}{S5} &
\multicolumn{2}{c}{Average} \\
\cmidrule(lr){2-3}\cmidrule(lr){4-5}\cmidrule(lr){6-7}\cmidrule(lr){8-9}\cmidrule(lr){10-11}\cmidrule(lr){12-13}
 & SR & AS & SR & AS & SR & AS & SR & AS & SR & AS & SR & AS \\
\midrule
DMRS-1D          & 0.500 & 17.4 & 0.625 & 15.8 & 0.625 & 18.3 & 0.750 & 15.1 & 0.500 & 19.3 & 0.600 & 17.2 \\
DMRS-2D          & 0.500 & 18.9 & 0.500 & 18.3 & 0.375 & 20.6 & 0.250 & 18.9 & 0.250 & 21.1 & 0.375 & 19.6 \\
CMRS             & 0.875 & 13.1 & 0.625 & 16.6 & 0.625 & 18.5 & 0.375 & 18.1 & 0.625 & 15.9 & 0.625 & 16.5 \\
Primitive MCTS   & 0.000 & 21.5 & 0.000 & 21.8 & 0.000 & 22.5 & 0.000 & 20.5 & 0.000 & 22.0 & 0.000 & 21.7 \\
LLM-MCTS         & 0.250 & 20.0 & 0.250 & 20.4 & 0.250 & 21.3 & 0.125 & 19.9 & 0.125 & 20.9 & 0.200 & 20.5 \\

COHERENT                     & \textbf{1.000} & 13.1 &
                               \textbf{1.000} & 11.4 &
                               \textbf{1.000} & 11.9 &
                               \textbf{1.000} & 11.4 &
                               \textbf{0.875} & 14.0 &
                               \textbf{0.975} & 12.4 \\

\midrule
\ourmethod (Ours)            & \textbf{1.000} & \textbf{10.8} &
                               \textbf{1.000} & \textbf{10.4} &
                               \textbf{1.000} & \textbf{11.8} &
                               \textbf{1.000} & \textbf{10.4} &
                               \textbf{0.875} & \textbf{11.4} &
                               \textbf{0.975} & \textbf{10.9} \\
Ground Truth (GT)            &   --  & 10.3 &   --  & 10.4 &   --  & 10.8 &   --  & 9.8  &   --  & 10.5 &   --  & 10.3 \\
\bottomrule
\end{tabular}
}
\caption{
    \textbf{Comparison Across Scenes in the COHERENT Benchmark.}
    \ourmethod{} outperforms all the baselines in every scene, demonstrating its superior performance. \\ 
}
\label{tab:coherent2}
\end{table*}

\begin{table*}[t]
\centering
\resizebox{0.85\linewidth}{!}{
\begin{tabular}{lcccccccccc}
\toprule
\multirow{2}{*}{Method} &
\multicolumn{2}{c}{Task 1 (Easy)} &
\multicolumn{2}{c}{Task 2 (Easy)} &
\multicolumn{2}{c}{Task 3 (Hard)} &
\multicolumn{2}{c}{Task 4 (Hard)} & 
\multicolumn{2}{c}{Average}\\
\cmidrule(lr){2-3}\cmidrule(lr){4-5}\cmidrule(lr){6-7}\cmidrule(lr){8-9}\cmidrule(lr){10-11}
 & SR & AS & SR & AS & SR & AS & SR & AS & SR & AS \\
\midrule
DMRS-1D   & 0.000 & 15.0 & 0.000 & 15.0 & 0.000 & 19.0 & 0.000 & 19.0 & 0.000 & 17.0\\
CMRS      & 0.000 & 15.0 & 0.000 & 15.0 & 0.000 & 19.0 & 0.000 & 19.0 & 0.000 & 17.0\\
COHERENT& \textbf{1.000} & 13.6 & 0.800 & 13.6& 0.400& 18.2& 0.600 & 17.8& 0.700& 15.8\\
\midrule
\ourmethod{} (Ours)  & \textbf{1.000} & \textbf{8.2} & \textbf{1.000} & \textbf{8.4} 
                     & \textbf{1.000} & \textbf{9.4} & \textbf{1.000} & \textbf{9.2}
                     & \textbf{1.000} & \textbf{8.8} \\
Ground Truth (GT) & -- & 7.0 & -- & 7.0 & -- & 9.0 & -- & 9.0 & -- & 8.0\\
\bottomrule
\end{tabular}
}
\caption{
    \textbf{Comparison Across Tasks in \ourbenchmark{}.}
    \ourmethod{} outperforms all our baselines and reduces the Average Step (AS) by over 40\%. \\
}
\label{tab:benchmark}
\end{table*}

\begin{table*}[t]
\centering

\resizebox{0.85\linewidth}{!}{
\begin{tabular}{lcccccccccc}
\toprule
\multirow{2}{*}{Method} &
\multicolumn{2}{c}{Task 1 (Easy)} &
\multicolumn{2}{c}{Task 2 (Easy)} &
\multicolumn{2}{c}{Task 3 (Hard)} &
\multicolumn{2}{c}{Task 4 (Hard)} & 
\multicolumn{2}{c}{Average}\\
\cmidrule(lr){2-3}\cmidrule(lr){4-5}\cmidrule(lr){6-7}\cmidrule(lr){8-9}\cmidrule(lr){10-11}
 & SR & AS & SR & AS & SR & AS & SR & AS  & SR & AS\\
 \midrule
\ourmethod{} w/o history  & 0.000 & 15.0 & 0.000 & 15.0 & 0.000 & 19.0 & 0.000 & 19.0 & 0.000 & 17.0 \\
\ourmethod{} w/o feedback & 0.200& 14.8& 0.200& 14.8& 0.200& 18.8& 0.200& 18.8& 0.200& 16.8\\
\ourmethod{} w/o grouping & 0.600& 14.0& 0.800 & 13.2& 0.600& 17.2& 0.600& 17.4& 0.650& 15.5\\
\midrule
\ourmethod{} (Ours)        & \textbf{1.000}& \textbf{8.2}
                           & \textbf{1.000}& \textbf{8.4}
                           & \textbf{1.000}& \textbf{9.4}
                           & \textbf{1.000}& \textbf{9.2}
                           & \textbf{1.000}& \textbf{8.8}\\
Ground Truth (GT)        &   --  & 7.0&   --  & 7.0&   --  & 9.0&   --  &  9.0& --&8.0\\
\bottomrule
\end{tabular}
}
\caption{
    \textbf{Ablation Studies.}
    Removing dialogue history, feedback information, or the grouping stage significantly reduces both Success Rate (SR) and Average Step (AS).\\
}
\label{tab:bench_ablation}
\end{table*}

In this section, we present a comprehensive evaluation of \ourmethod{} to address the following questions:
\begin{enumerate}[label=(\arabic*)]
\setlength{\itemindent}{0pt}
\setlength{\itemsep}{0em}
\item \label{q:one} Is \ourmethod{} effective for simple daily-life multi-robot collaboration tasks?
\item \label{q:two} Can \ourmethod{} perform well in challenging multi-robot assembly tasks?
\item \label{q:three} Through ablation studies, how critical are the individual components of \ourmethod{}?
\end{enumerate}

We evaluate \ourmethod{} in two distinct environments: \ourbenchmark{} and a simpler heterogeneous multi-robot collaboration benchmark from COHERENT~\cite{liu2025coherent}.
For LLM use, we use \textit{gpt-4-0125-preview} to align with the setting in COHERENT.
For quantitative analysis, we use task Success Rate (SR) and Average Step (AS) as evaluation metrics in this paper.

\subsection{Evaluating {\ourmethod} on Simple Daily-Life Multi-Robot Collaboration}

To answer Question~\ref{q:one}, we evaluate \ourmethod{} on the COHERENT benchmark, a simpler heterogeneous multi-robot benchmark 
that includes diverse tasks across five real-world scenes, but involves at most three heterogeneous robots and assumes perfect skill execution.  
We adopt its evaluation metrics and use the reported results as our baseline.

Results shown in Table~\ref{tab:coherent1} and~\ref{tab:coherent2} suggest that \ourmethod{} succeeds on nearly all COHERENT tasks and achieves higher efficiency with fewer steps.  
This trend holds across every scene, demonstrating our \ourmethod{}' superior performance.  
Notably, in the most challenging trio-type tasks, which require all three robots to collaborate, \ourmethod{} delivers the largest gain, reducing the Average Step count by 18.6\%,  
indicating that our approach offers stronger improvements on more complex tasks.

\subsection{Evaluating \ourmethod{} on \ourbenchmark{} with Robot Assembly Tasks}
\label{sec:exp_main_evaluate}

To answer Question~\ref{q:two}, we evaluate \ourmethod{} on \ourbenchmark{}.
Our baselines include the following:
\begin{itemize}
\setlength{\itemindent}{0pt}
\setlength{\leftskip}{-10pt}
\setlength{\itemsep}{0em}
    \item DMRS-1D: a variant of CoELA~\cite{zhang2024building}, this decentralized framework lets robots determine their next step through dialogue, with the final decision summarized by the last robot.
    \item CMRS: a primitive centralized system~\cite{huang2022language} that uses a single decision-making LLM to output executable actions, where all information is stored in the prompt.
    \item COHERENT: an approximately centralized baseline that combines an oracle planning LLM with a feedback LLM, where dialogue history is used to adjust future perception and planning.
\end{itemize}

For quantitative evaluation, we fix the scene parameters and select four representative scenarios to evaluate all methods.
For each scenario, we manually derive minimal-step solutions as ground-truth references.
We observe that scenarios with heavier object occlusion require more ground-truth steps, naturally forming two levels of difficulty.

Due to stochastic skill execution in \ourbenchmark{}, each task is executed five times, and we report the mean Success Rate (SR) and Average Steps (AS).
To control evaluation cost and runtime, a task is considered successful only if it is completed within twice the corresponding ground-truth step count.


Results in Table~\ref{tab:benchmark} show that \ourmethod{} achieves a 100\% success rate on \ourbenchmark{}, outperforming all baselines.
In addition, \ourmethod{} reduces the Average Steps (AS) by 44.30\% compared to the best baseline, indicating substantially higher planning efficiency in long-horizon collaboration.

Notably, most baselines terminate due to timeout and fail to complete tasks within the step budget, and the performance degradation of baselines from the COHERENT benchmark (Table~\ref{tab:coherent1}) to \ourbenchmark{} (Table~\ref{tab:benchmark}) demonstrates that \ourbenchmark{} provides a significantly more challenging evaluation setting for robot collaboration.

\textbf{Recover from Execution Failures.}
From our experiments, we observe three types of execution failures:
(1) \emph{improper grouping}, where no robot in the assigned group can complete the subtask;
(2) \emph{incorrect agent selection}, where a valid subtask is assigned to a robot without the required capabilities; and
(3) \emph{state inconsistency}, where missing information or unmet preconditions prevent proper execution.
Nevertheless, the natural-language feedback provided by the executors captures these failure signals and, through the context memory, enables downstream perception and planning to be refined, ultimately leading to successful task completion.


\subsection{Ablation Studies on \ourmethod{}}

To answer Question~\ref{q:three}, we conduct an ablation study to assess the contribution of each component in \ourmethod{} by:
(i) removing the dialogue history,
(ii) removing the feedback information, and
(iii) removing the grouping stage from the grouping–planning–execution–feedback cycle.
We evaluate all variants on the same tasks and metrics as in Section~\ref{sec:exp_main_evaluate}, with results summarized in Table~\ref{tab:bench_ablation}.

The results show that both dialogue history and feedback information are critical to the framework: removing either component leads to failure in task completion, indicating that sustained context accumulation and execution-aware feedback are indispensable for long-horizon planning.
The grouping stage also plays a central role in \ourmethod{}, significantly improving task success rates and reducing average steps by enabling effective coordination and parallelization across heterogeneous robots.

\section{Conclusion}
In this paper, we present \ourmethod{}, a human-team-inspired adaptive group negotiation framework among multiple LLMs for heterogeneous multi-robot collaboration, together with \ourbenchmark{}, a challenging benchmark for long-horizon assembly tasks under realistic execution uncertainty.
Extensive experiments show that \ourmethod{} consistently outperforms all baselines in both success rate and efficiency, while robustly recovering from execution failures with feedback-driven negotiation.
Ablation studies highlight the complementary roles of dialogue history, feedback, and dynamic subgroup formation.
Overall, our results demonstrate that human-inspired group formation and negotiation principles offer an effective and scalable solution for heterogeneous multi-robot collaboration.

\section{Limitations}
In this paper, we primarily focus on improving the efficiency of multi-robot collaboration at the level of high-level behavioral planning.
We assume that LLM inference latency is negligible compared to skill execution time, and therefore do not explicitly model network delays or computational costs in the current framework.
Managing API costs under inference-efficiency constraints and exploring asynchronous inference–execution pipelines remain important directions for future work.
In addition, while \ourbenchmark{} emphasizes long-horizon behavioral planning and coordination, it simplifies sensory inputs and perception modules in simulation.
Bridging this gap toward more realistic perception and enabling sim-to-real transfer constitute promising avenues for future research.

\section{Ethical Considerations}
Our study investigates multi-robot collaboration using large language models (LLMs) for planning and negotiation in simulated environments.
We discuss the primary ethical considerations relevant to this work below.

\begin{itemize}
\setlength{\itemindent}{0pt}
\setlength{\leftskip}{-10pt}
\setlength{\itemsep}{0em}

\item \textbf{No Human or Sensitive Data.}
Our research is conducted entirely in simulation and does not involve human subjects, personally identifiable information, or sensitive real-world data.
We do not collect or process user data, nor do we rely on proprietary datasets.

\item \textbf{Safety and Responsible Deployment.}
While our methods and benchmark are evaluated exclusively in simulation, real-world deployment of autonomous multi-robot systems may pose physical safety risks.
In particular, LLM-based planning may be susceptible to erroneous or inconsistent decisions under distribution shift.
Any future deployment should therefore incorporate rigorous safety validation, fail-safe mechanisms, and human oversight, and comply with applicable safety standards and regulations.

\item \textbf{Bias and Model Limitations.}
The LLMs used in our study are pretrained by third parties and may reflect societal biases present in their training data.
Although our work does not directly deploy LLMs in human-facing decision-making, such biases could indirectly influence planning behavior.
We acknowledge this limitation and recommend bias auditing and robustness testing before any real-world application.

\end{itemize}

\bibliography{main}

@String(NIPS= {Adv. Neural Inform. Process. Syst.})

@String(AAAI = {AAAI})

@String(NIPS  = {NeurIPS})

@misc{liu2024leveraginglargelanguagemodel,
      title={Leveraging Large Language Model for Heterogeneous Ad Hoc Teamwork Collaboration}, 
      author={Xinzhu Liu and Peiyan Li and Wenju Yang and Di Guo and Huaping Liu},
      year={2024},
      eprint={2406.12224},
      archivePrefix={arXiv},
      primaryClass={cs.RO},
      url={https://arxiv.org/abs/2406.12224}, 
}

@article{wu2023tidybot,
  title = {TidyBot: Personalized Robot Assistance with Large Language Models},
  author = {Wu, Jimmy and Antonova, Rika and Kan, Adam and Lepert, Marion and Zeng, Andy and Song, Shuran and Bohg, Jeannette and Rusinkiewicz, Szymon and Funkhouser, Thomas},
  journal = {Autonomous Robots},
  year = {2023}
}

@article{wang2023voyager,
  title   = {Voyager: An Open-Ended Embodied Agent with Large Language Models},
  author  = {Guanzhi Wang and Yuqi Xie and Yunfan Jiang and Ajay Mandlekar and Chaowei Xiao and Yuke Zhu and Linxi Fan and Anima Anandkumar},
  year    = {2023},
  journal = {arXiv preprint arXiv: Arxiv-2305.16291}
}

@INPROCEEDINGS{progprompt,
  author={Singh, Ishika and Blukis, Valts and Mousavian, Arsalan and Goyal, Ankit and Xu, Danfei and Tremblay, Jonathan and Fox, Dieter and Thomason, Jesse and Garg, Animesh},
  booktitle={2023 IEEE International Conference on Robotics and Automation (ICRA)}, 
  title={ProgPrompt: Generating Situated Robot Task Plans using Large Language Models}, 
  year={2023},
  volume={},
  number={},
  pages={11523-11530},
  doi={10.1109/ICRA48891.2023.10161317}}

@misc{brown2020fewshot,
      title={Language Models are Few-Shot Learners}, 
      author={Tom B. Brown and Benjamin Mann and Nick Ryder and Melanie Subbiah and Jared Kaplan and Prafulla Dhariwal and Arvind Neelakantan and Pranav Shyam and Girish Sastry and Amanda Askell and Sandhini Agarwal and Ariel Herbert-Voss and Gretchen Krueger and Tom Henighan and Rewon Child and Aditya Ramesh and Daniel M. Ziegler and Jeffrey Wu and Clemens Winter and Christopher Hesse and Mark Chen and Eric Sigler and Mateusz Litwin and Scott Gray and Benjamin Chess and Jack Clark and Christopher Berner and Sam McCandlish and Alec Radford and Ilya Sutskever and Dario Amodei},
      year={2020},
      eprint={2005.14165},
      archivePrefix={arXiv},
      primaryClass={cs.CL},
      url={https://arxiv.org/abs/2005.14165}, 
}

@article{huang2022language,
      title={Language Models as Zero-Shot Planners: Extracting Actionable Knowledge for Embodied Agents},
      author={Huang, Wenlong and Abbeel, Pieter and Pathak, Deepak and Mordatch, Igor},
      journal={arXiv preprint arXiv:2201.07207},
      year={2022}
    }

@inproceedings{llmzeroreason,
author = {Kojima, Takeshi and Gu, Shixiang Shane and Reid, Machel and Matsuo, Yutaka and Iwasawa, Yusuke},
title = {Large language models are zero-shot reasoners},
year = {2022},
isbn = {9781713871088},
publisher = {Curran Associates Inc.},
address = {Red Hook, NY, USA},
booktitle = {Proceedings of the 36th International Conference on Neural Information Processing Systems},
articleno = {1613},
numpages = {15},
location = {New Orleans, LA, USA},
series = {NIPS '22}
}

@misc{mandi2023roco,
	title={RoCo: Dialectic Multi-Robot Collaboration with Large Language Models}, 
	author={Zhao Mandi and Shreeya Jain and Shuran Song},
	year={2023},
	eprint={2307.04738},
	archivePrefix={arXiv},
	primaryClass={cs.RO}
}

@article{chen2023reconcile,
  title={ReConcile: Round-Table Conference Improves Reasoning via Consensus Among Diverse LLMs},
  author={Chen, Justin Chih-Yao and Saha, Swarnadeep and Bansal, Mohit},
  journal={arXiv preprint arXiv:2309.13007},
  year={2023}
}

@inproceedings{hong2024metagpt,
      title={Meta{GPT}: Meta Programming for A Multi-Agent Collaborative Framework},
      author={Sirui Hong and Mingchen Zhuge and Jonathan Chen and Xiawu Zheng and Yuheng Cheng and Jinlin Wang and Ceyao Zhang and Zili Wang and Steven Ka Shing Yau and Zijuan Lin and Liyang Zhou and Chenyu Ran and Lingfeng Xiao and Chenglin Wu and J{\"u}rgen Schmidhuber},
      booktitle={The Twelfth International Conference on Learning Representations},
      year={2024},
      url={https://openreview.net/forum?id=VtmBAGCN7o}
}

@article{liang2023encouraging,
  title={Encouraging Divergent Thinking in Large Language Models through Multi-Agent Debate},
  author={Liang, Tian and He, Zhiwei and Jiao, Wenxiang and Wang, Xing and Wang, Yan and Wang, Rui and Yang, Yujiu and Tu, Zhaopeng and Shi, Shuming},
  journal={arXiv preprint arXiv:2305.19118},
  year={2023}
}

@misc{zhang2024building,
      title={Building Cooperative Embodied Agents Modularly with Large Language Models}, 
      author={Hongxin Zhang and Weihua Du and Jiaming Shan and Qinhong Zhou and Yilun Du and Joshua B. Tenenbaum and Tianmin Shu and Chuang Gan},
      year={2024},
      eprint={2307.02485},
      archivePrefix={arXiv},
      primaryClass={cs.AI},
      url={https://arxiv.org/abs/2307.02485}, 
}

@article{kannan2023smart,
 	title={SMART-LLM: Smart Multi-Agent Robot Task Planning using Large Language Models},
  	author={Kannan, Shyam Sundar and Venkatesh, Vishnunandan LN and Min, Byung-Cheol},
  	journal={arXiv preprint arXiv:2309.10062},
 	year={2023}
}

@inproceedings{liu2025coherent,
  title={Coherent: Collaboration of heterogeneous multi-robot system with large language models},
  author={Liu, Kehui and Tang, Zixin and Wang, Dong and Wang, Zhigang and Li, Xuelong and Zhao, Bin},
  booktitle={2025 IEEE International Conference on Robotics and Automation (ICRA)},
  pages={10208--10214},
  year={2025},
  organization={IEEE}
}

@article{chen2023scalable,
  title={Scalable Multi-Robot Collaboration with Large Language Models: Centralized or Decentralized Systems?},
  author={Chen, Yongchao and Arkin, Jacob and Zhang, Yang and Roy, Nicholas and Fan, Chuchu},
  journal={arXiv preprint arXiv:2309.15943},
  year={2023}
}

@inproceedings{saycan,
    title={Do As I Can and Not As I Say: Grounding Language in Robotic Affordances},
    author={Michael Ahn and Anthony Brohan and Noah Brown and Yevgen Chebotar and Omar Cortes and Byron David and Chelsea Finn and Chuyuan Fu and Keerthana Gopalakrishnan and Karol Hausman and Alex Herzog and Daniel Ho and Jasmine Hsu and Julian Ibarz and Brian Ichter and Alex Irpan and Eric Jang and Rosario Jauregui Ruano and Kyle Jeffrey and Sally Jesmonth and Nikhil Joshi and Ryan Julian and Dmitry Kalashnikov and Yuheng Kuang and Kuang-Huei Lee and Sergey Levine and Yao Lu and Linda Luu and Carolina Parada and Peter Pastor and Jornell Quiambao and Kanishka Rao and Jarek Rettinghouse and Diego Reyes and Pierre Sermanet and Nicolas Sievers and Clayton Tan and Alexander Toshev and Vincent Vanhoucke and Fei Xia and Ted Xiao and Peng Xu and Sichun Xu and Mengyuan Yan and Andy Zeng},
    booktitle={arXiv preprint arXiv:2204.01691},
    year={2022}
}

@article{text2motion,
  title={Text2Motion: from natural language instructions to feasible plans},
  author={Lin, Kevin and Agia, Christopher and Migimatsu, Toki and Pavone, Marco and Bohg, Jeannette},
  journal={Autonomous Robots},
  year={2023},
  month={Nov},
  day={14},
  issn={1573-7527},
  doi={10.1007/s10514-023-10131-7},
  url={https://doi.org/10.1007/s10514-023-10131-7}
}

@inproceedings{codeaspolicies,
  author={Liang, Jacky and Huang, Wenlong and Xia, Fei and Xu, Peng and Hausman, Karol and Ichter, Brian and Florence, Pete and Zeng, Andy},
  booktitle={2023 IEEE International Conference on Robotics and Automation (ICRA)}, 
  title={Code as Policies: Language Model Programs for Embodied Control}, 
  year={2023},
  volume={},
  number={},
  pages={9493-9500},
  doi={10.1109/ICRA48891.2023.10160591}
}

@article{kashyap2021particle,
  title={Particle Swarm Optimization aided PID gait controller design for a humanoid robot},
  author={Kashyap, Abhishek Kumar and Parhi, Dayal R},
  journal={ISA transactions},
  volume={114},
  pages={306--330},
  year={2021},
  publisher={Elsevier}
}

@article{katayama2023model,
  title={Model predictive control of legged and humanoid robots: models and algorithms},
  author={Katayama, Sotaro and Murooka, Masaki and Tazaki, Yuichi},
  journal={Advanced Robotics},
  volume={37},
  number={5},
  pages={298--315},
  year={2023},
  publisher={Taylor \& Francis}
}

@misc{narang2022factoryfastcontactrobotic,
      title={Factory: Fast Contact for Robotic Assembly}, 
      author={Yashraj Narang and Kier Storey and Iretiayo Akinola and Miles Macklin and Philipp Reist and Lukasz Wawrzyniak and Yunrong Guo and Adam Moravanszky and Gavriel State and Michelle Lu and Ankur Handa and Dieter Fox},
      year={2022},
      eprint={2205.03532},
      archivePrefix={arXiv},
      primaryClass={cs.RO},
      url={https://arxiv.org/abs/2205.03532}, 
}

@inproceedings{gao2024coohoi,
 author = {Gao, Jiawei and Wang, Ziqin and Xiao, Zeqi and Wang, Jingbo and Wang, Tai and Cao, Jinkun and Hu, Xiaolin and Liu, Si and Dai, Jifeng and Pang, Jiangmiao},
 booktitle = {Advances in Neural Information Processing Systems},
 title = {CooHOI: Learning Cooperative Human-Object Interaction with Manipulated Object Dynamics},
 year = {2024}
}

@article{Peng_2021,
   title={AMP: adversarial motion priors for stylized physics-based character control},
   volume={40},
   ISSN={1557-7368},
   url={http://dx.doi.org/10.1145/3450626.3459670},
   DOI={10.1145/3450626.3459670},
   number={4},
   journal={ACM Transactions on Graphics},
   publisher={Association for Computing Machinery (ACM)},
   author={Peng, Xue Bin and Ma, Ze and Abbeel, Pieter and Levine, Sergey and Kanazawa, Angjoo},
   year={2021},
   month=jul, pages={1–20} }

@article{rajeswaran2017learning,
  title={Learning complex dexterous manipulation with deep reinforcement learning and demonstrations},
  author={Rajeswaran, Aravind and Kumar, Vikash and Gupta, Abhishek and Vezzani, Giulia and Schulman, John and Todorov, Emanuel and Levine, Sergey},
  journal={arXiv preprint arXiv:1709.10087},
  year={2017}
}

@article{li2025reinforcement,
  title={Reinforcement learning for versatile, dynamic, and robust bipedal locomotion control},
  author={Li, Zhongyu and Peng, Xue Bin and Abbeel, Pieter and Levine, Sergey and Berseth, Glen and Sreenath, Koushil},
  journal={The International Journal of Robotics Research},
  volume={44},
  number={5},
  pages={840--888},
  year={2025},
  publisher={SAGE Publications Sage UK: London, England}
}

@inproceedings{bellegarda2024visual,
  title={Visual CPG-RL: Learning central pattern generators for visually-guided quadruped locomotion},
  author={Bellegarda, Guillaume and Shafiee, Milad and Ijspeert, Auke},
  booktitle={2024 IEEE International Conference on Robotics and Automation (ICRA)},
  pages={1420--1427},
  year={2024},
  organization={IEEE}
}

@article{perolat2022mastering,
  title={Mastering the game of Stratego with model-free multiagent reinforcement learning},
  author={Perolat, Julien and De Vylder, Bart and Hennes, Daniel and Tarassov, Eugene and Strub, Florian and de Boer, Vincent and Muller, Paul and Connor, Jerome T and Burch, Neil and Anthony, Thomas and others},
  journal={Science},
  volume={378},
  number={6623},
  pages={990--996},
  year={2022},
  publisher={American Association for the Advancement of Science}
}

@inproceedings{kurach2020google,
  title={Google research football: A novel reinforcement learning environment},
  author={Kurach, Karol and Raichuk, Anton and Sta{\'n}czyk, Piotr and Zaj{\k{a}}c, Micha{\l} and Bachem, Olivier and Espeholt, Lasse and Riquelme, Carlos and Vincent, Damien and Michalski, Marcin and Bousquet, Olivier and others},
  booktitle={Proceedings of the AAAI conference on artificial intelligence},
  volume={34},
  number={04},
  pages={4501--4510},
  year={2020}
}

@article{lai2025roboballet,
  title={RoboBallet: Planning for multirobot reaching with graph neural networks and reinforcement learning},
  author={Lai, Matthew and Go, Keegan and Li, Zhibin and Kr{\"o}ger, Torsten and Schaal, Stefan and Allen, Kelsey and Scholz, Jonathan},
  journal={Science Robotics},
  volume={10},
  number={106},
  pages={eads1204},
  year={2025},
  publisher={American Association for the Advancement of Science}
}

@inproceedings{wang2024multi,
  title={Multi-robot cooperative socially-aware navigation using multi-agent reinforcement learning},
  author={Wang, Weizheng and Mao, Le and Wang, Ruiqi and Min, Byung-Cheol},
  booktitle={2024 IEEE International Conference on Robotics and Automation (ICRA)},
  pages={12353--12360},
  year={2024},
  organization={IEEE}
}

@article{li2022v2x,
  title={V2X-Sim: Multi-agent collaborative perception dataset and benchmark for autonomous driving},
  author={Li, Yiming and Ma, Dekun and An, Ziyan and Wang, Zixun and Zhong, Yiqi and Chen, Siheng and Feng, Chen},
  journal={IEEE Robotics and Automation Letters},
  volume={7},
  number={4},
  pages={10914--10921},
  year={2022},
  publisher={IEEE}
}

@inproceedings{yang2018mean,
  title={Mean field multi-agent reinforcement learning},
  author={Yang, Yaodong and Luo, Rui and Li, Minne and Zhou, Ming and Zhang, Weinan and Wang, Jun},
  booktitle={International conference on machine learning},
  pages={5571--5580},
  year={2018},
  organization={PMLR}
}

@inproceedings{rein2024gpqa,
  title={Gpqa: A graduate-level google-proof q\&a benchmark},
  author={Rein, David and Hou, Betty Li and Stickland, Asa Cooper and Petty, Jackson and Pang, Richard Yuanzhe and Dirani, Julien and Michael, Julian and Bowman, Samuel R},
  booktitle={First Conference on Language Modeling},
  year={2024}
}

@article{jain2024livecodebench,
  title={Livecodebench: Holistic and contamination free evaluation of large language models for code},
  author={Jain, Naman and Han, King and Gu, Alex and Li, Wen-Ding and Yan, Fanjia and Zhang, Tianjun and Wang, Sida and Solar-Lezama, Armando and Sen, Koushik and Stoica, Ion},
  journal={arXiv preprint arXiv:2403.07974},
  year={2024}
}

@article{plaat2024reasoning,
  title={Reasoning with large language models, a survey},
  author={Plaat, Aske and Wong, Annie and Verberne, Suzan and Broekens, Joost and van Stein, Niki and B{\"a}ck, Thomas},
  journal={CoRR},
  year={2024}
}

@inproceedings{song2023llm,
  title={Llm-planner: Few-shot grounded planning for embodied agents with large language models},
  author={Song, Chan Hee and Wu, Jiaman and Washington, Clayton and Sadler, Brian M and Chao, Wei-Lun and Su, Yu},
  booktitle={Proceedings of the IEEE/CVF international conference on computer vision},
  pages={2998--3009},
  year={2023}
}

@article{zhang2024badrobot,
  title={Badrobot: Jailbreaking llm-based embodied ai in the physical world},
  author={Zhang, Hangtao and Zhu, Chenyu and Wang, Xianlong and Zhou, Ziqi and Hu, Shengshan and Zhang, Leo Yu},
  journal={arXiv preprint arXiv:2407.20242},
  volume={3},
  year={2024}
}

@article{mower2024ros,
  title={Ros-llm: A ros framework for embodied ai with task feedback and structured reasoning},
  author={Mower, Christopher E and Wan, Yuhui and Yu, Hongzhan and Grosnit, Antoine and Gonzalez-Billandon, Jonas and Zimmer, Matthieu and Wang, Jinlong and Zhang, Xinyu and Zhao, Yao and Zhai, Anbang and others},
  journal={arXiv preprint arXiv:2406.19741},
  year={2024}
}

@article{salimpour2025towards,
  title={Towards Embodied Agentic AI: Review and Classification of LLM-and VLM-Driven Robot Autonomy and Interaction},
  author={Salimpour, Sahar and Fu, Lei and Keramat, Farhad and Militano, Leonardo and Toffetti, Giovanni and Edelman, Harry and Queralta, Jorge Pe{\~n}a},
  journal={arXiv preprint arXiv:2508.05294},
  year={2025}
}

@article{liang2025large,
  title={Large Model Empowered Embodied AI: A Survey on Decision-Making and Embodied Learning},
  author={Liang, Wenlong and Zhou, Rui and Ma, Yang and Zhang, Bing and Li, Songlin and Liao, Yijia and Kuang, Ping},
  journal={arXiv preprint arXiv:2508.10399},
  year={2025}
}

@article{liu2024leveraging,
  title={Leveraging large language model for heterogeneous ad hoc teamwork collaboration},
  author={Liu, Xinzhu and Li, Peiyan and Yang, Wenju and Guo, Di and Liu, Huaping},
  journal={arXiv preprint arXiv:2406.12224},
  year={2024}
}

@inproceedings{mandi2024roco,
  title={Roco: Dialectic multi-robot collaboration with large language models},
  author={Mandi, Zhao and Jain, Shreeya and Song, Shuran},
  booktitle={2024 IEEE International Conference on Robotics and Automation (ICRA)},
  pages={286--299},
  year={2024},
  organization={IEEE}
}

@article{makoviychuk2021isaac,
  title={Isaac gym: High performance gpu-based physics simulation for robot learning},
  author={Makoviychuk, Viktor and Wawrzyniak, Lukasz and Guo, Yunrong and Lu, Michelle and Storey, Kier and Macklin, Miles and Hoeller, David and Rudin, Nikita and Allshire, Arthur and Handa, Ankur and others},
  journal={arXiv preprint arXiv:2108.10470},
  year={2021}
}

@inproceedings{mahmood2019amass,
  title={AMASS: Archive of motion capture as surface shapes},
  author={Mahmood, Naureen and Ghorbani, Nima and Troje, Nikolaus F and Pons-Moll, Gerard and Black, Michael J},
  booktitle={Proceedings of the IEEE/CVF international conference on computer vision},
  pages={5442--5451},
  year={2019}
}

@article{schulman2017proximal,
  title={Proximal policy optimization algorithms},
  author={Schulman, John and Wolski, Filip and Dhariwal, Prafulla and Radford, Alec and Klimov, Oleg},
  journal={arXiv preprint arXiv:1707.06347},
  year={2017}
}

\newpage
\appendix
{
\noindent\Large\textbf{Appendix}
}






\section{Additional Details on \ourbenchmark{}}
\label{sec:benchmark details}

\subsection{Simulation Setup and Agent Parameters}
\label{sec:benchmark_sim_agents}

\textbf{Physics Simulation.}
All tasks in \ourbenchmark{} are developed in the IsaacGym physics simulator~\cite{makoviychuk2021isaac}.
The simulator executes robot actions in a step-by-step manner, advancing the physical state of the environment after each issued skill.
Gravity, collision detection, and joint limit constraints are enabled for all agents to ensure physically consistent execution.

\textbf{Agent Roster and Physical Specifications.}
The assembly scene in \ourbenchmark{} includes three types of heterogeneous robotic agents:
a single robotic manipulator, three autonomous ground vehicles (AGVs), and one humanoid.
Table~\ref{tab:agentparameters} summarizes the key specifications of each agent type, including degrees of freedom, payload capacity, maximum reach, and the number of instantiated agents.

\begin{table}[h]
   \centering
\resizebox{1.0\linewidth}{!}{
    \begin{tabular}{ccccc}\toprule
         &  DoF &  Payload & Max Reach & Quantity \\\midrule
         Franka & 7 &  3kg & 855mm & 1 \\
         TRACER Mini &  3 &  80kg & All & 3 \\
         Humanoid &  28 & -- & All & 1 \\ \bottomrule
    \end{tabular}
}
\caption{\textbf{Agent Parameters in {\ourbenchmark}.}}
\label{tab:agentparameters}
\end{table}

The robotic arm is instantiated as a Franka Panda manipulator with seven degrees of freedom and a maximum reach of 855\,mm, suitable for precise manipulation and assembly.
The AGVs are instantiated as TRACER Mini mobile robots with differential-drive kinematics and high payload capacity, enabling reliable transportation of components within the workspace.
The humanoid features a high-dimensional articulated body with 28 degrees of freedom, supporting whole-body locomotion and object carrying skills.


\subsection{Environment Setup and Randomization}
\label{sec:benchmark_env_init}

\textbf{Environment Workspace Initialization.}
All tasks in \ourbenchmark{} are executed within a shared global coordinate frame.
The workspace domain, component placement candidates, and agent initialization ranges are summarized in Table~\ref{tab:initialization}.
All object poses and agent base positions are specified in this global frame, while the vertical axis is used only to model object height and collision geometry.

\begin{table}[h]
\centering
\resizebox{1.0\linewidth}{!}{
\begin{tabular}{ll}
\toprule
Entity & Initialization Range \\
\midrule
Environment Domain & $(x, y) \in [-6, 6] \times [-10, 10]$ \\
Components (Wheels \& Trunk) & $(x, y) \in \{(4,8),(-4,8),(-4,-8),(4,-8)\}$ \\
AGVs & $x \in [-3,3], \; y \in [-5,5]$ \\
Humanoid  & Central region of the environment \\
\bottomrule
\end{tabular}
}
\caption{\textbf{Random Initialization} of the Environment Components and Agents in \ourbenchmark{}.}
\label{tab:initialization}
\end{table}

\textbf{Obstacle Configuration.}
In addition to randomly initialized agents and components, the environment includes static obstacles that block navigation and cause occlusions.
Four rectangular obstacles of size $1 \times 5\,\mathrm{m}$ are placed near the corners of the workspace.
Three smaller obstacles of size $1 \times 1 \times 0.5\,\mathrm{m}$ are positioned along primary traversal paths of the mobile agents, introducing potential collisions and blocked access regions.

\textbf{Randomization Policy.}
For each episode, component locations and agent initial positions are independently sampled from the ranges specified in Table~\ref{tab:initialization}.
All methods are evaluated under identical environment configurations and random seeds to ensure fair comparison.
No curriculum learning or adaptive environment scheduling is employed.

\subsection{Low-level Skill Implementation Details}
\label{sec:benchmark_skill_impl}


\textbf{Franka Manipulation Implementation.}
Franka manipulation skills are executed using a numerical inverse kinematics (IK) solver combined with an operational-space controller (OSC).
The IK solver computes target joint configurations for the first seven joints, with a maximum of 200 iterations and a residual tolerance of $10^{-4}$.
The Franka base is initialized at a fixed pose $(0, -2.0, 0)$ with orientation quaternion $(0, 0, 0, 1)$.
Task-space motion is controlled using an impedance-based OSC with:
\begin{equation}
\mathbf{F} = k_p \, \mathbf{e} + k_v \, \dot{\mathbf{e}},
\end{equation}
where the proportional gain is set to $k_p = 5$ and the derivative gain is chosen as $k_v = 2\sqrt{k_p}$ to achieve critical damping.
The gripper closing threshold is set to $0.05\,\mathrm{m}$, accounting for the object grasping radius of $0.03\,\mathrm{m}$ used in \ourbenchmark{}.

During assembly, components are aligned and attached using a magnetic attraction model with an effective range of $0.03\,\mathrm{m}$.
For each manipulation trajectory, operational-space waypoints are generated via interpolation to ensure smooth and continuous execution.

\textbf{AGV Navigation and Motion Control.}
AGV transportation skills are implemented using a Rapidly-exploring Random Tree (RRT) planner operating in the planar workspace.
The planner considers all static and dynamic obstacles and robots in the environment as the collision space.
The RRT planner uses a collision-checking resolution of $0.1\,\mathrm{m}$ along each edge and a rewire count of 32.
Once a feasible path is found, it is executed using differential-drive control to enable smooth turning and stable navigation.
For final positioning during delivery, the AGV follows a straight-line motion segment to improve placement reliability.

\textbf{Humanoid Observation and Skill Interface.}
Humanoid skills are executed by a pretrained policy that operates on a structured observation interface.
Let $N_j = 15$ denote the number of articulated bodies of the humanoid.
For each body $j \in \{1, \dots, N_j\}$, the observation is defined as
\begin{equation}
\mathbf{o}_j = \big[\mathbf{q}_j, \mathbf{p}_j, \mathbf{v}_j, \mathbf{a}_j \big],
\end{equation}
where $\mathbf{q}_j \in \mathbb{R}^6$ represents the body relative rotational and translational configuration,
$\mathbf{p}_j \in \mathbb{R}^3$ the Cartesian position,
$\mathbf{v}_j \in \mathbb{R}^3$ the linear velocity, and
$\mathbf{a}_j \in \mathbb{R}^3$ the linear acceleration.
The full humanoid observation vector is constructed by concatenating all body-level observations and a task observation:
\begin{equation}
\mathbf{o} = [\mathbf{o}_1, \mathbf{o}_2, \dots, \mathbf{o}_{N_j}, \mathbf{o}_{task}].
\end{equation}
Observations of the root body and global reference frame are excluded, as they encode absolute pose information rather than local kinematic and dynamic states.
This formulation provides a consistent interface for skill execution while remaining independent of the humanoid’s global position and orientation in the environment.

\subsection{Humanoid Carry Skill Training Details}
\label{sec:benchmark_humanoid_training}

\textbf{AMP Objective and Reward Design.}
Humanoid carrying skills are trained using the Adversarial Motion Priors (AMP) framework~\cite{Peng_2021}, following the single-object manipulation paradigm introduced in COOHOI~\cite{gao2024coohoi}.
AMP augments the task objective with a discriminator-based style reward that encourages motions consistent with human motion data.
At each time step $t$, the total reward is defined as
\begin{equation}
r_t = w^{G} r^{G}(\mathbf{s}_t, \mathbf{g}_t, \mathbf{s}_{t+1}) + w^{S} r^{S}(\mathbf{s}_t, \mathbf{s}_{t+1}),
\end{equation}
where $r^{G}$ denotes the task-specific reward and $r^{S}$ is the AMP style reward.
The task reward consists of components for walking, holding, and placement:
\begin{equation}
r^{G} = 0.2\, r^{G}_{\text{walk}} + 0.4\, r^{G}_{\text{held}} + 0.4\, r^{G}_{\text{target}},
\label{eq:task}
\end{equation}
while the style reward is computed as
\begin{equation}
r^{S}(\mathbf{s}_t, \mathbf{s}_{t+1}) = -\log\big(1 - D(\mathbf{s}_t, \mathbf{s}_{t+1})\big),
\label{eq:style}
\end{equation}
with $D(\cdot)$ denoting the discriminator trained on reference human motion data.

\begin{table}[h]
\centering
\resizebox{0.6\linewidth}{!}{
\begin{tabular}{cc}
\hline
Parameters                               & Value \\ \hline
num envs                                 & 16384          \\
episode length                           & 600            \\
discount factor                          & 0.99           \\
e\_clip                                  & 0.2            \\
horizon\_length                          & 32             \\
minibatch\_size                          & 16384          \\
\multicolumn{1}{l}{amp\_minibatch\_size} & 4096           \\
\multicolumn{1}{l}{task\_reward\_weight} & 0.5            \\
\multicolumn{1}{l}{disc\_reward\_weight} & 0.5            \\ \hline
\end{tabular}
}
\caption{\textbf{Humanoid Carrying Skill Training.}}
\label{tab:AMP training}
\end{table}

\textbf{Discriminator Training.}
The AMP discriminator is trained to distinguish between policy-generated motion transitions and reference human motion data.
We adopt the same discriminator architecture and training procedure as in~\cite{Peng_2021}, where the discriminator operates on pairs of consecutive states $(\mathbf{s}_t, \mathbf{s}_{t+1})$ and outputs the probability that the transition originates from real motion data.
Reference motions are drawn from the ACCAD subset of the AMASS dataset~\cite{mahmood2019amass}, which contains a diverse set of human motions including locomotion, lifting, carrying, and object placement.
During training, the discriminator and the policy are updated alternately.
The discriminator is trained using binary cross-entropy loss to separate reference transitions from policy-generated transitions, while the policy receives the discriminator-based style reward defined in Eq.~\ref{eq:style}.
This adversarial training scheme encourages the humanoid policy to produce stable and human-like motions while optimizing task-specific objectives.

\begin{figure*}[h]
  \centering
   \includegraphics[width=0.9\linewidth]{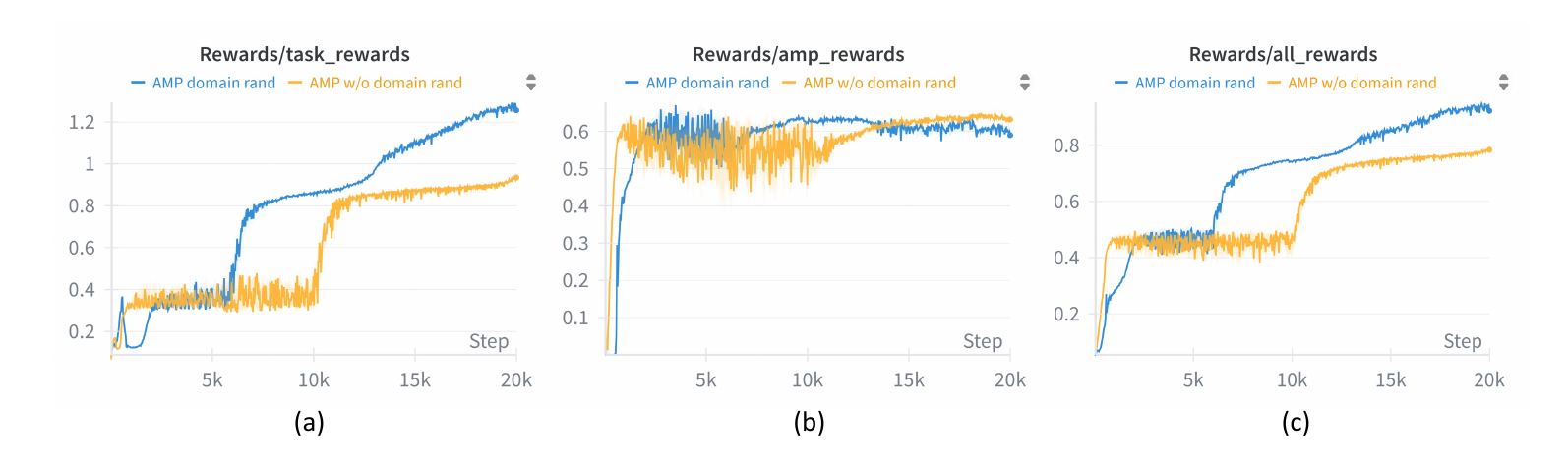}
   \caption{\textbf{Ablation Study on AMP Domain Randomization.} Results show that Domain Randomization helps the humanoid agent to obtain a stable and reusable carrying skill in \ourbenchmark{}.
}
\label{fig:rl}
\end{figure*}

\begin{figure*}
  \centering
   \includegraphics[width=1.0\linewidth]{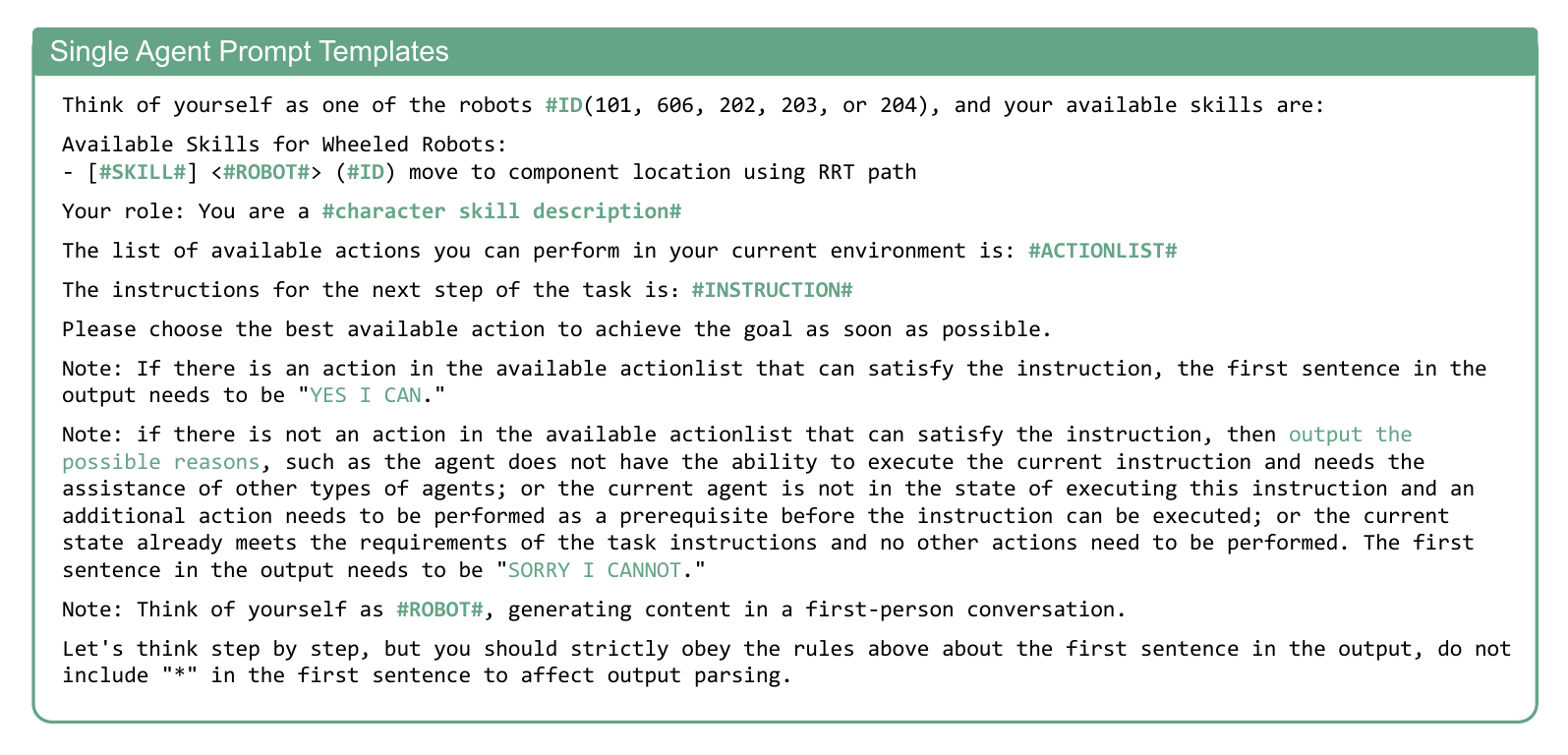}
   \caption{\textbf{Single Agent Prompt Templates.} This template encodes the agent’s name, unique identifier, and action specification to ensure structured and unambiguous command interpretation for an \emph{agent executor}.
}
\label{fig:action_ins}
\end{figure*}

\textbf{Domain Randomization.}
To improve robustness and generalization in assembly environments, we apply domain randomization during training.
Randomized factors include object orientation, relative alignment between the humanoid and the object, and object proximity.
These variations encourage the learned policy to tolerate spatial uncertainty and contact perturbations commonly encountered during execution in \ourbenchmark{}.

\textbf{Training Hyperparameters.}
The humanoid policy is trained using Proximal Policy Optimization (PPO)~\cite{schulman2017proximal}.
Key training hyperparameters, including the number of parallel environments, episode length, reward weights, and optimization settings, are summarized in Table~\ref{tab:AMP training}.

\textbf{Training Infrastructure.}
Training is conducted in IsaacGym headless mode on a single NVIDIA A100 GPU to enable large-scale parallel simulation.
After training, the learned policy is deployed with rendering enabled on a workstation equipped with an NVIDIA RTX~4060~Ti for evaluation.

\textbf{Ablation on Domain Randomization.}
Fig.~\ref{fig:rl} presents an ablation on domain randomization in AMP training.
Without randomization, it converges to a local optimum early and exhibits poor generalization, while domain-randomized training achieves consistently higher task rewards and more reliable execution.
This result supports the use of domain randomization to obtain a stable and reusable carrying skill for the benchmark.

\subsection{LLM-to-Skill Execution Interface}
\label{sec:benchmark_llm_interface}

\textbf{Command Schema and Prompt Templates.}
All robot actions in \ourbenchmark{} are issued through structured natural-language commands generated by LLM-based planners.
Each command explicitly specifies the action type, the acting agent(s), the target object, and the target location.
Figure~\ref{fig:action_ins} illustrates the single-agent prompt templates used to ensure parsable command generation.
These templates encode the agent name, unique identifier, and action specification, providing a unified interface across heterogeneous robots.

\begin{figure*}
  \centering
   \includegraphics[width=0.8\linewidth]{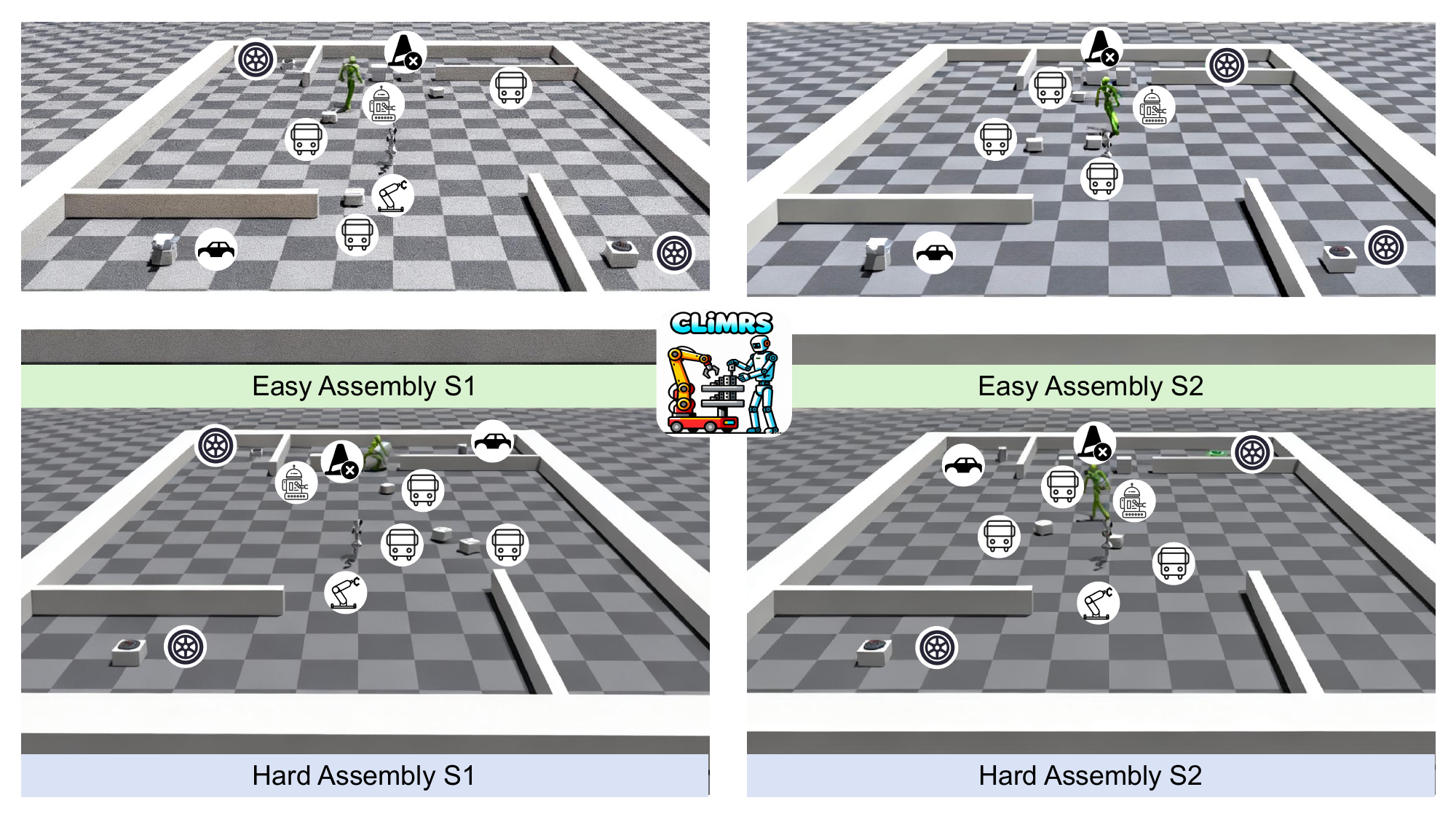}
   \caption{\textbf{Tasks of Different Difficulties.} To evaluate the effectiveness of our method in facilitating multi-robot collaboration, we design four heterogeneous agent scenarios under two difficulty tiers, namely \textit{easy} and \textit{hard}. 
}
\label{fig:env}
\end{figure*}

\textbf{Command Parsing via Regular Expressions.}
Given an LLM-generated command $C$, the execution interface parses it into a structured representation using a set of predefined regular expressions.
Formally, each command is mapped as
\begin{equation}
C \xrightarrow{\text{parse}} \{G, A, O, L\},
\end{equation}
where $G$ denotes the group identifier, $A = \{(a_i, \text{id}_i)\}$ is the set of assigned agents with their identifiers, $O$ specifies the target object and its identifier, and $L$ denotes the target location.
Only commands that can be successfully parsed into this structure are considered for further verification.


\textbf{Command Verification.}
To ensure reliable and safe execution, each parsed command is validated through a four-stage verification pipeline.
Let $f$ denote an LLM function call corresponding to a structured command.
The verification operator $V$ maps $f$ to a validation vector
\begin{equation}
f \xrightarrow{\;V\;} \{v_1, v_2, v_3, v_4\},
\end{equation}
where each component corresponds to a specific verification stage.
The pipeline consists of the following four stages:

\begin{figure*} [h]
  \centering
   \includegraphics[width=0.85\linewidth]{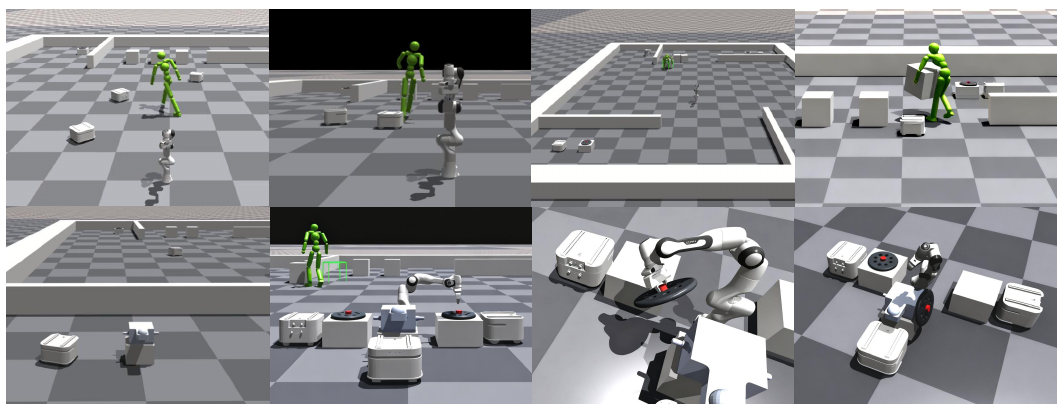}
   \caption{\textbf{Simulation.} It shows an example of the assembly process in the dynamic simulation.
}
\label{fig:env2}
\end{figure*}

\begin{figure*}
  \centering
   \includegraphics[width=1.0\linewidth]{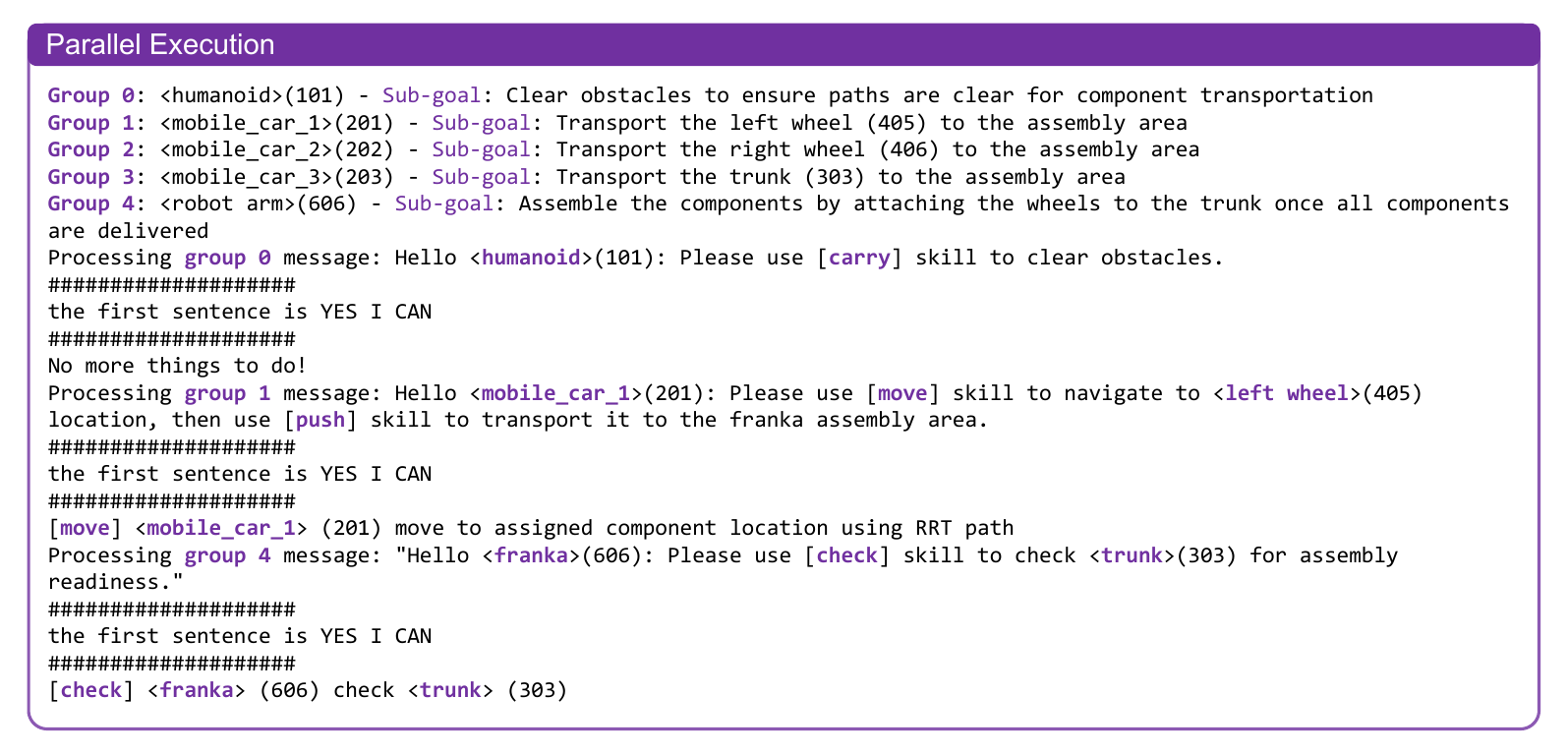}
   \caption{\textbf{Parallel Execution.} To enhance task execution efficiency, agents within each group are executed in parallel, as reflected in the execution log.
}
\label{fig:parallel}
\end{figure*}

\begin{itemize}
\setlength{\itemindent}{0pt}
\setlength{\leftskip}{-10pt}
\setlength{\itemsep}{0em}
  \item \textbf{Capability assessment.}
  The system first evaluates whether the assigned agent(s) are capable of executing the requested action based on their available skills.
  The assessment yields a binary decision or confidence score, and commands below a preset threshold $\tau_c$ are rejected or deferred.

  \item \textbf{Action selection.}
  For commands that pass the capability assessment, the planner selects a concrete action $a^\ast$ from the available action set $\mathcal{A}$, optionally providing a ranked subset or per-action confidence.

  \item \textbf{Action parsing.}
  The selected action $a^\ast$ is processed by a parsing operator $\mathcal{P}$ to produce a structured command
  $C^{\mathrm{struct}} = \mathcal{P}(a^\ast)$,
  which includes fields such as action type, agent assignment, target object, and target location.

  \item \textbf{Execution verification (judge).}
  A dedicated judge module $\mathcal{J}$ evaluates $C^{\mathrm{struct}}$ for format correctness, semantic consistency, physical feasibility, and execution safety.
\end{itemize}

Only commands that satisfy all verification criteria are forwarded to the execution module.
Overall, the verification operator can be expressed as
\begin{equation}
V = \mathcal{J} \circ \mathcal{P} \circ \mathcal{S}_2 \circ \mathcal{S}_1,
\end{equation}
where $\mathcal{S}_1$ and $\mathcal{S}_2$ denote the capability assessment and action selection stages, respectively.

\textbf{Failure Handling and Feedback Logging.}
Commands that fail at any stage of the verification pipeline are not executed.
Instead, the interface records structured failure feedback indicating the reason for rejection, such as insufficient agent capability, semantic mismatch, or physical infeasibility.
This feedback is propagated to the future planning process and stored in the \emph{context memory} module, enabling planners to adjust subsequent decisions.
Deferred commands may be reconsidered in later planning iterations once agent states change or environment conditions are ready.



\subsection{Benchmark Evaluation Details}
\label{sec:benchmark_protocol}

\textbf{Task Difficulty Settings.}
We define two levels of task difficulty, referred to as \emph{easy} and \emph{hard}, which differ in obstacle configuration and agent initialization.
In the easy setting, AGV transportation paths are free of blocking obstacles, and initial agent placements are arranged to avoid potential collisions with the humanoid.
In contrast, the hard setting introduces obstacles along AGV trajectories and initializes agents in closer proximity, increasing the likelihood of collisions and requiring more coordination during execution.
Examples of the two settings are illustrated in Fig.~\ref{fig:env}.

\textbf{Trials and Stochasticity.}
Due to stochasticity in low-level skill execution and environment interactions, each task configuration is evaluated over multiple independent trials.
Unless otherwise specified, we run each task five times and report averaged metrics across trials.
All methods are evaluated under identical environment configurations and random seeds to ensure fair comparison.
An example of the assembly process can be seen in Fig.~\ref{fig:env2}.

\textbf{Step Budget and Timeout Criteria.}
To control evaluation cost and ensure comparability, each task is assigned a maximum step budget.
The step budget is defined as twice the number of steps in a manually derived minimal-step solution for the corresponding task.
A task is considered unsuccessful if the target assembly is not completed within this budget, even if individual skills execute successfully.
This criterion penalizes inefficient long-horizon planning and excessive replanning.

\textbf{Execution Logging.}
During evaluation, the benchmark records detailed execution logs, including agent group assignments, issued skill commands, verification outcomes, and environment feedback at each step.
These logs enable post-hoc analysis of planning behavior, grouping dynamics, and failure recovery, and are used for qualitative visualization and debugging. An example of the parallel execution logs can be seen in Fig.~\ref{fig:parallel}.

\section{Example Logs and Prompts}
\label{sec:log}

We show some examples of logs and prompts in all the figures below.

\begin{figure*}
  \centering
   \includegraphics[width=1.0\linewidth]{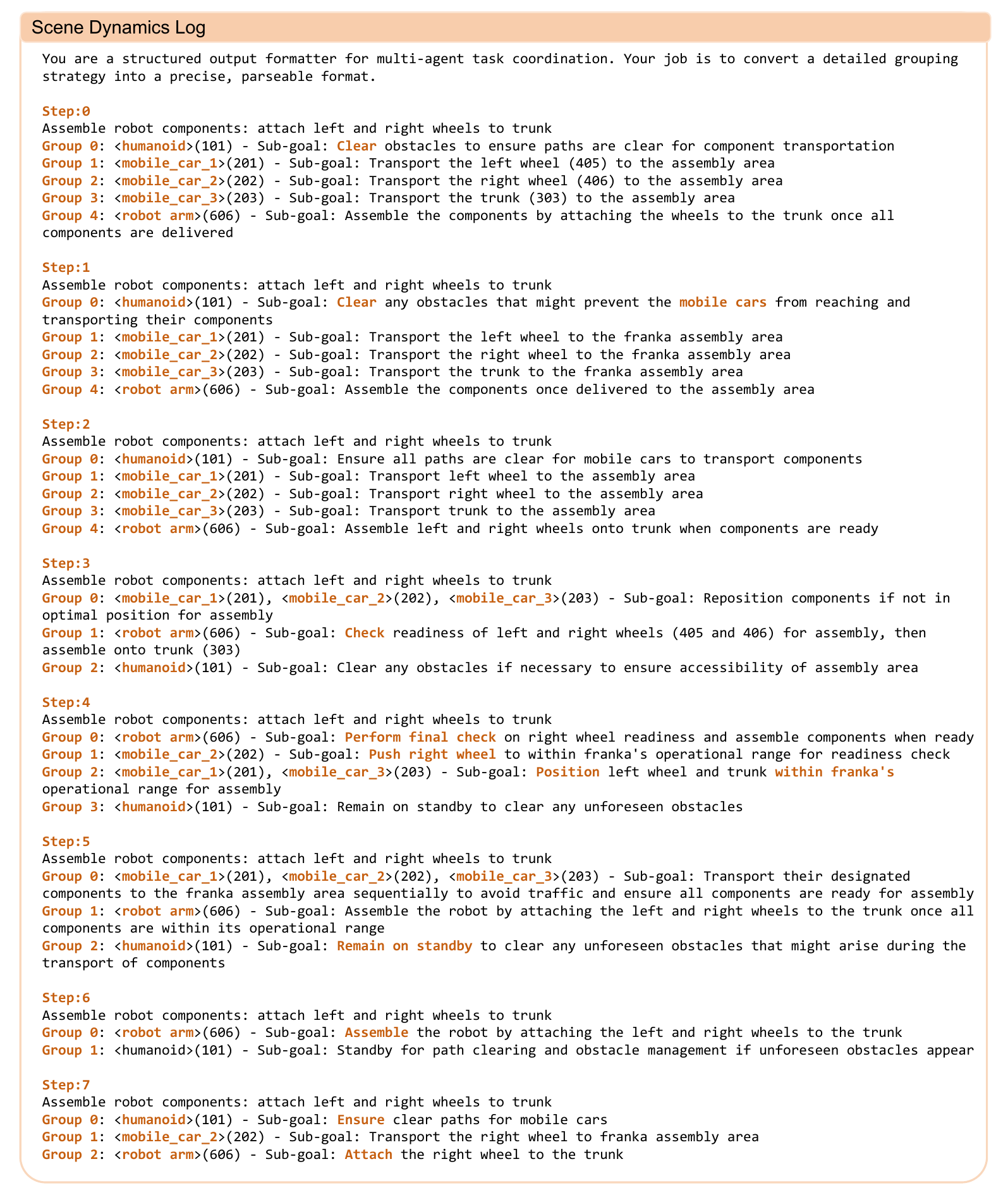}
   \caption{\textbf{Scene Dynamics Log.} It illustrates the formation of sub-groups and their corresponding sub-tasks.
   }
\label{fig:log}
\end{figure*}

\begin{figure*}
  \centering
   \includegraphics[width=1.0\linewidth]{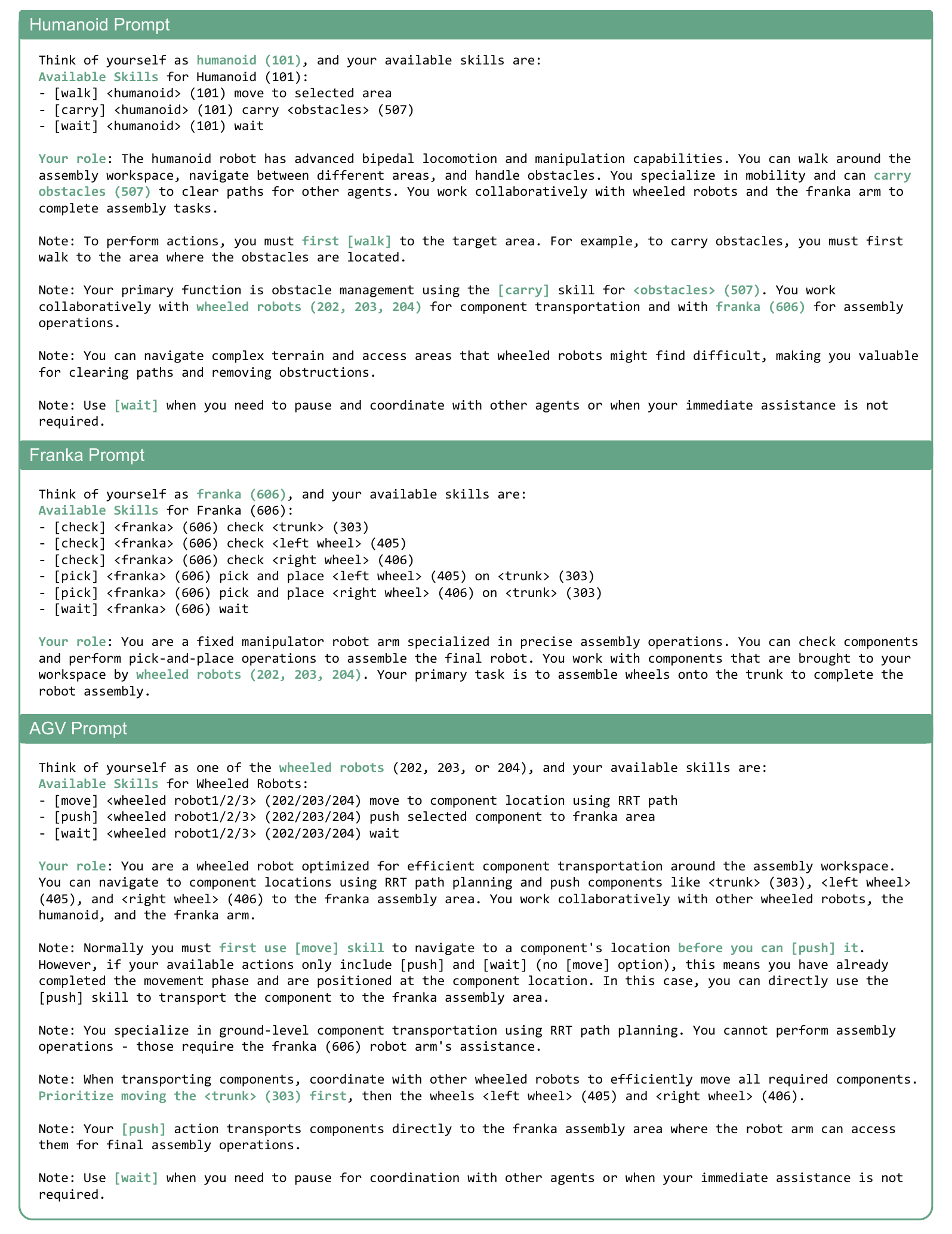}
   \caption{\textbf{Single Agent Prompts.} It illustrates the single-agent prompt design for Franka, Humanoid, and AGV, specifying their roles and available skills.
}
\label{fig:prompt_single_agent}
\end{figure*}

\begin{figure*}
  \centering
   \includegraphics[width=1.0\linewidth]{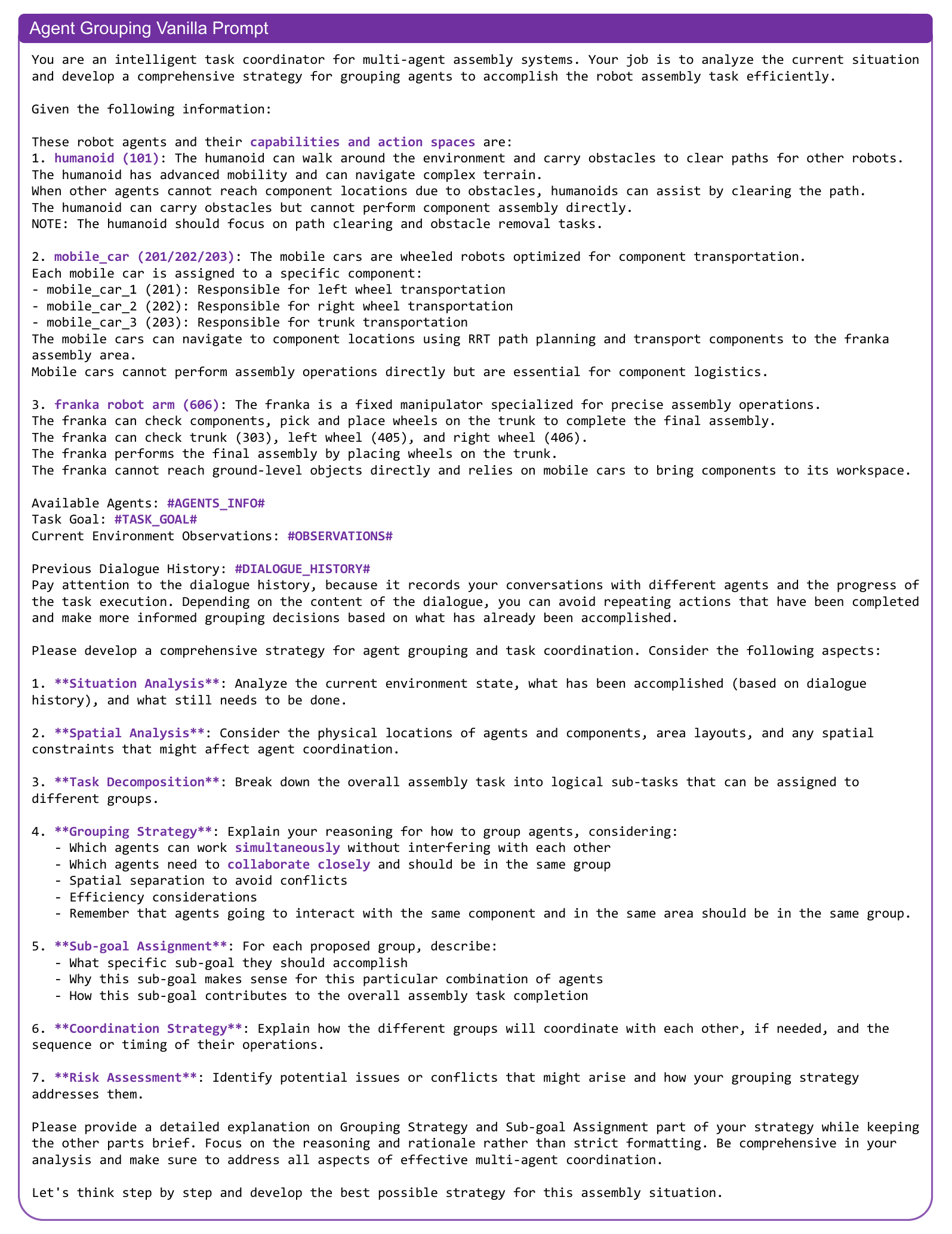}
   \caption{\textbf{Vanilla Agent Grouping Prompt.} It illustrates the prompt to extract agent sub-groups and their corresponding sub-tasks with the general proposal planner.}
\label{fig:prompt_vanilla_grouping}
\end{figure*}

\begin{figure*}
  \centering
   \includegraphics[width=1.0\linewidth]{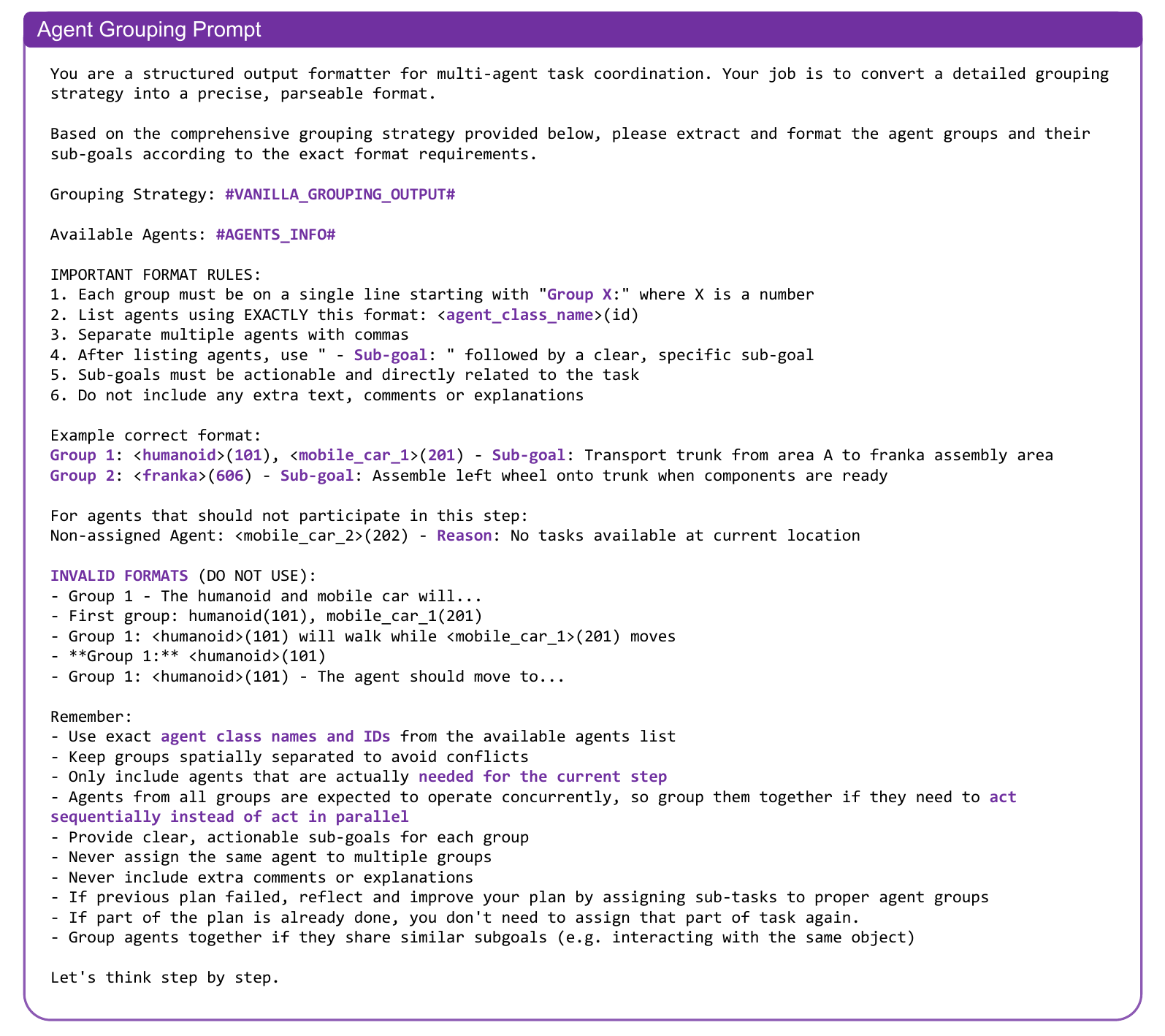}
   \caption{\textbf{Agent Grouping Prompt.} It illustrates the prompt to form agent sub-groups with a general proposal with the general proposal planner.
}
\label{fig:prompt_grouping}
\end{figure*}

\end{document}